\documentclass[10pt,twocolumn,letterpaper]{article}

\usepackage{cvpr}              % To produce the CAMERA-READY version
\definecolor{cvprblue}{rgb}{0.21,0.49,0.74}
\usepackage[pagebackref,breaklinks,colorlinks,allcolors=cvprblue]{hyperref}
\usepackage{array} % required for text wrapping in tables
\usepackage{algorithm}
\usepackage{algpseudocode}
\usepackage{amsmath, amssymb}
\usepackage[table]{xcolor}
\usepackage{newfloat}
\usepackage{listings}
\usepackage{hyperref}
\usepackage{color}
\usepackage{xspace}
\usepackage{booktabs}
\usepackage{makecell}
\usepackage{graphicx} 
\usepackage{bibentry}
\usepackage{xspace}
\usepackage{marvosym}
\usepackage{multirow}
\usepackage[accsupp]{axessibility}  % Improves PDF readability for those with disabilities.
\def\paperID{414} % *** Enter the Paper ID here
\def\confName{CVPR}
\def\confYear{2026}

\newcommand{\ModelNameUpper}{Task-aware Virtual View Exploration}

\newcommand{\ModelAbbr}{TVVE}

\usepackage{fontawesome}

\title{Learning to See and Act: Task-Aware Virtual View Exploration for \\Robotic Manipulation}

\author{
Yongjie Bai$^{1,2,*}$\quad
Zhouxia Wang$^{1,3,*}$\quad
Yang Liu$^{1,\textsuperscript{\Letter}}$ \quad
Kaijun Luo$^{1}$ \quad
Yifan Wen$^{1}$ \quad
Mingtong Dai$^{2,4}$ \\
Weixing Chen$^{1}$ \quad
Ziliang Chen$^{2}$\quad
Lingbo Liu$^{2}$\quad
Guanbin Li$^{1,2}$\quad
Liang Lin$^{1,2,5}$
\\
\small
$^1$Sun Yat-sen University 
$^2$Pengcheng Laboratory
$^3$Nanyang Technological University \\
\small
$^4$Shenzhen Institutes of Advanced Technology, Chinese Academy of Sciences
$^5$X-Era AI Lab 
}

\begin{document}
\maketitle

\let\thefootnote\relax\footnotetext{$^*$ Equal Contribution\hspace{3pt} ~\faEnvelopeO ~Corresponding author\hspace{5pt} \\ Accepted by CVPR 2026
}

% \begingroup
% \renewcommand{\thefootnote}{}
% \footnotetext{$^*$ Interns in ARC Lab, Tencent PCG\hspace{3pt} ~\faEnvelopeO ~Corresponding author}
% \endgroup

% \footnotetext[$^*$]{Interns in ARC Lab, Tencent PCG. Corresponding author.}

\begin{abstract}

Recent vision-language-action (VLA) models for multi-task robot manipulation often rely on fixed camera setups and shared visual encoders, which limit their performance under occlusions and during cross-task transfer. To address these challenges, we propose \textbf{Task-aware Virtual View Exploration (TVVE)}, a framework that learns to select task-relevant \emph{virtual} camera viewpoints and dynamically re-render observations from a reconstructed scene representation using the selected viewpoints. To enable efficient view selection, we train an exploration policy in a \emph{pseudo-environment}. In addition, we introduce a \textbf{Task-aware Mixture-of-Experts (TaskMoE)} visual encoder that routes visual features to task-specialized experts, mitigating interference in multi-task learning. 
To evaluate robustness under distribution shifts, we construct \textbf{RLBench-OG}, an out-of-distribution benchmark with visual perturbations and camera pose variations. Experiments on RLBench and RLBench-OG demonstrate that TVVE achieves higher success rates than strong baselines, while real-robot experiments further confirm its robustness to visual disturbances and unseen instructions. Code and visualizations are available at: \href{https://hcplab-sysu.github.io/TAVP}{TAVP}.

\end{abstract}    
\section{Introduction}
\label{sec:into}
A general-purpose robotic system is expected to accurately perform numerous manipulation tasks \cite{liu2025aligning}. With the development of vision-language-action (VLA) models \cite{kimopenvla,rt-1, rt-2,livision, cheang2024gr, bu2024towards, li2025bridgevla, bu2025univla, cen2025worldvla, liu2025hybridvla, team2024octo, lin2025onetwovla} and transformer-based architectures \cite{ACT, rt-1, rt-2, hou2024diffusion, chi2023diffusion}, many end-to-end robotic manipulation frameworks (\eg, OpenVLA\cite{kimopenvla}, $\pi_0$\cite{black2410pi0}) have achieved promising results on  complex and fine-grained tasks. Typically, these models rely on observations captured from a single view \cite{kimopenvla, chi2023diffusion, shridhar2022cliport} or a few fixed viewpoints \cite{ke20243ddiffactor, shridhar2023perceiver} to guide action prediction. While this may suffice in simple environments with only a few objects, it becomes problematic in cluttered or dynamic scenes—particularly when objects or the end-effector are repositioned during task execution.

\begin{figure}
    \centering
    \includegraphics[scale=0.42]{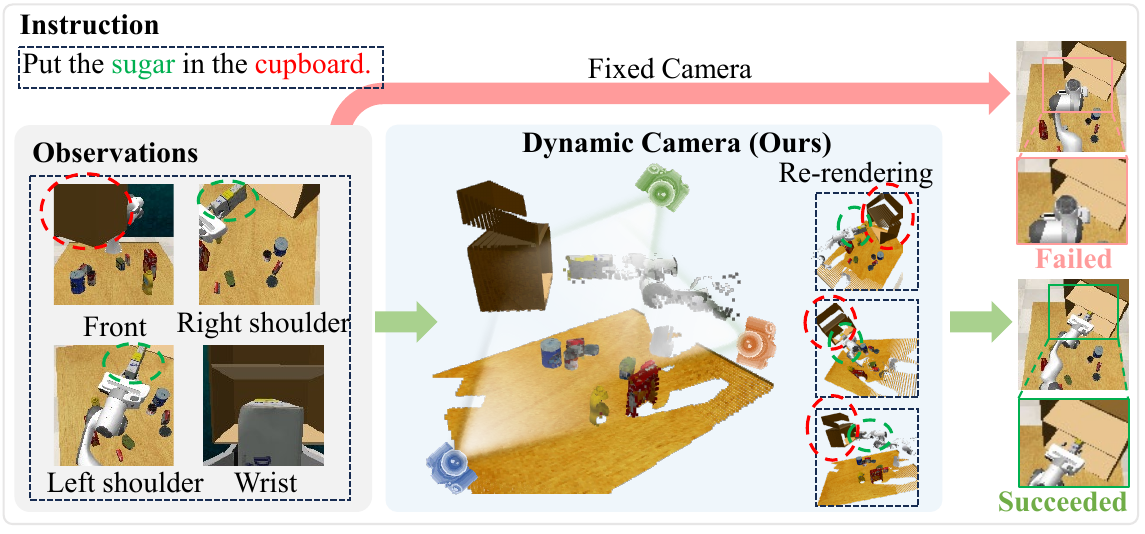}
    \vspace{-10pt}
    \caption{\textbf{Motivation Illustration.} Observations captured from fixed cameras often miss parts of the target objects. For example, the front view only captures the cupboard (highlighted with a \textcolor{red}{red circles}), while the left and right shoulder views only show the sugar (already grasped by the end-effector and highlighted with \textcolor{green}{green circles}). These incomplete observations may lead to failed operations. In contrast, our proposed \ModelAbbr\ is designed to dynamically explore and re-render informative viewpoints that maximize coverage of target-relevant information, thereby improving the reliability of manipulation outcomes.}
       \vspace{-15pt}
    \label{fig:motivation}
\end{figure}

In such scenarios, fixed viewpoints often lead to occlusions of the target object or the end-effector, resulting in incomplete scene understanding and inaccurate action predictions. As shown in Fig.~\ref{fig:motivation}, given the instruction \textit{``Put the sugar in the cupboard"}, images captured from three fixed cameras either miss the cupboard or fail to include the end-effector grasped with sugar. Such incomplete visual observations prevent the robot from correctly interpreting the scene, ultimately leading to suboptimal or failed actions.

Beyond acquiring complete and informative visual features, a scalable and task-sensitive visual encoder is also critical for accurate perception in multi-task settings. Although prior language-conditioned transformer-based methods, such as RVT~\cite{goyal2023rvt} and RVT-2~\cite{goyal2024rvt2}, enable multi-task capabilities by attending to instruction tokens, their shared encoders still suffer from task interference, especially when handling visually and semantically diverse tasks (\eg, \textit{pick an apple} vs. \textit{open a drawer}). This entanglement limits the generalization and scalability of such models across a broad range of manipulation behaviors.

To address these challenges, we propose the \textbf{\ModelNameUpper\ (\ModelAbbr)} model, which integrates viewpoint exploration with task-aware visual feature extraction for robust and generalizable multi-task robotic manipulation. \ModelAbbr\ explores informative viewpoints using a delicately designed efficient reinforcement learning–based exploration policy, while employing a scalable task-aware Mixture-of-Experts (MoE) visual encoder to extract task-specific features. This design enables the model to obtain more complete and discriminative visual information, thereby improving robustness and generalization in multi-task action prediction.

Specifically, our task-aware Mixture-of-Experts module, termed~\textbf{TaskMoE}, improves multi-task accuracy and generalization through two key designs:
(1) a dynamic expert routing mechanism guided by fused instruction and scene cues, rather than relying solely on task identifiers; and
(2) a decoupled gating strategy that enables parameter sharing among semantically similar tasks, while isolating routing for semantically diverse ones.
Based on the extracted task-specific features, we introduce the Multi-Viewpoint Exploration Policy (MVEP) that adaptively selects task-specific camera poses based on the global point cloud. By re-rendering 2D observations from the explored viewpoints, MVEP enriches visual perception with globally aware features, leading to more informed action prediction and improved manipulation success.

% to train a viewpoint exploration policy that selects optimal camera poses for each task. The Policy explores the optimal camera re-rendering viewpoints based on the global point cloud. Then, it renders the corresponding 2D images from these viewpoints for subsequent feature extraction and action prediction. These features possess enhanced \textit{global awareness} with richer information, capable of improving robotic performance.

To comprehensively evaluate the effectiveness of our \ModelAbbr\ in enhancing robotic spatial understanding and action generation performance, as well as its multi-task perception and decision-making capabilities, we conducted rigorous benchmark testing on 18 diverse RLBench~\cite{james2020rlbench} simulation tasks covering multiple tasks and scenarios. Our contributions are summarized as follows:
\begin{itemize}
    \item \textbf{Multi-Viewpoint Exploration Policy (MVEP) :} The MVEP enables robots to explore better virtual observation viewpoints (dynamic multi-view re-rendering), effectively addressing occlusion and insufficient viewpoint problems, and enhancing 3D perception capabilities.
    \item \textbf{Task-aware Mixture-of-Experts (TaskMoE):} We introduce TaskMoE, which dynamically selects perception and action generation experts according to the task and its related instruction and scene visual information, improving multi-task processing capability and robustness.
    \item \textbf{RLBench-OG:} We construct a new benchmark to comprehensively evaluate the model robustness and generalization under out-of-distribution (OOD) settings, covering visual perturbations such as occlusions, background changes, lighting variations, texture shifts, and camera pose changes.
    \item \textbf{Superior Robotic Manipulation Performance:} Extensive experiments on RLBench, RLBench-OG and real-robot tasks demonstrate that our \ModelAbbr\ significantly outperforms existing baselines in multi-task manipulation w.r.t accuracy and robustness.
\end{itemize}

\section{Related Works}
\label{sec:related}
\subsection{Multi-task Learning in Manipulation}

The core challenges for general-purpose robots lie in task generalization capabilities and multi-task execution in complex environments. In the past, multi-task learning \cite{deisenroth2014multi, ze2023gnfactor, goyal2024rvt2, ma2024hierarchical} has achieved remarkable progress. It primarily consists of two approaches: Modular solutions decompose skills into reusable basic units (e.g., perception, planning, and control modules) \cite{mao2024robomatrix}. Though interpretable, they face the combinatorial explosion problem—as task complexity increases, the cost of designing interfaces and coordinating between modules grows exponentially, leading to system rigidity. For instance, ManipGen \cite{dalallocal} consumed massive training resources to train over 1,000 task-specific experts. Conversely, end-to-end solutions employ a single dense model to directly map perception to action \cite{team2024octo, ze20243d, wang2024poco, chi2023diffusion, shridhar2022cliport, nair2022learning, haldar2024baku}. While simplifying the pipeline, significant differences in visual feature distributions and motion patterns across tasks (e.g., grasping fragile objects vs. tightening screws) cause the model to struggle with convergence during training due to parameter conflicts. The Mixture of Experts (MoE) \cite{shazeer2017outrageously} addresses the scaling bottleneck of large models through a sparse activation mechanism. %Its core idea is to partition the model into multiple expert sub-networks (Experts) and introduce a Gating Network to dynamically route input data. 
SDP \cite{wangsparse} integrated MoE into diffusion policies to address multi-task learning and task transfer. Differently, we propose a task-aware MoE (TaskMoE) to resolve feature conflicts between different robotic manipulation tasks, activating distinct task-specific experts for different tasks, thereby enhancing multi-task execution performance and generalization capabilities.

\begin{figure*}[t]
\centering
\includegraphics[scale=0.45]{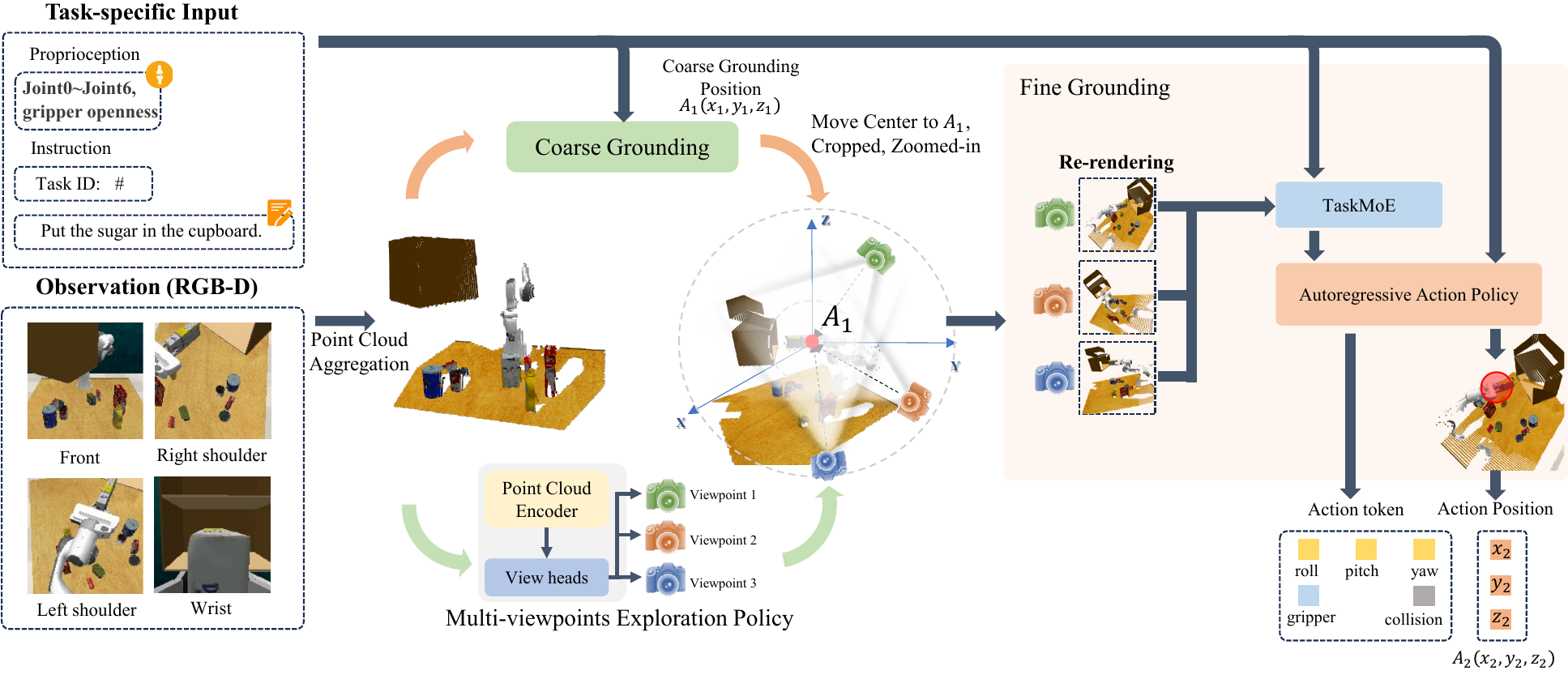}
\vspace{-12pt}
\caption{The overview of our \ModelNameUpper\ (\ModelAbbr) framework. The input of this framework is multiple RGB-D images from fixed viewpoints. First, it converts them into point clouds and aggregates them into a global point cloud in the world coordinate system. Then, it diverges into two branches: One branch \textcolor{orange}{(orange)} performs \textbf{Coarse Grounding} to predict the approximate position of the end-effector. Subsequently, it moves the center of the global point cloud to this predicted position, performs scaling and cropping, retaining the important point cloud region. Another branch \textcolor{green}{(green)} receives the global point cloud, passes it through MVEP to predict the optimal camera parameters for the observation viewpoint. Then, using these parameters, it renders a 2D image from the point cloud processed by the red branch. This rendered image is fed into the \textbf{Fine Grounding} to predict the final robot action, including the end-effector position, rotation, gripper status, and collision state.}
\vspace{-15pt}
\label{fig:framework}
\end{figure*}

\subsection{Vision-Language-Action Model}

Recently, numerous VLA models \cite{shridhar2022cliport, team2024octo, ze20243d, goyal2024rvt2, zhang2025autoregressive} have been proposed. %These models take linguistic instructions and visual observations as input and directly output low-level execution actions. 
%Unlike traditional approaches involving pose estimation \cite{fang2020graspnet, deng2020self} and path planning \cite{prianto2020path, zhao2025deep}, this significantly streamlines the model pipeline. 
In contrast to traditional pipelines that rely on explicit pose estimation~\cite{fang2020graspnet, deng2020self} and path planning~\cite{prianto2020path, zhao2025deep}, VLA models directly map linguistic instructions and visual observations to low-level actions, substantially simplifying the control stack. With the advancement of large language models (LLMs) and vision-language models (VLMs), these large models have been applied in robotics \cite{kimopenvla, rt-1, rt-2, livision, cheang2024gr, bu2024towards}, leveraging their powerful priors to provide world knowledge for robots, thereby enhancing model generalization and performance. However, most existing VLA models operate on a single or fixed viewpoint, limiting a robot’s ability to form comprehensive 3D spatial understanding and thereby constraining action prediction. To overcome this limitation, we introduce a dynamic multi-view re-rendering perception scheme that enhances spatial perception and scene comprehension, ultimately improving execution robustness in complex environments.

\subsection{Reinforcement Learning in Robotics}

Reinforcement learning (RL) has achieved significant accomplishments in robotics \cite{tang2025deep}. However, implementing RL starting from scratch typically necessitates complex reward mechanism and training paradigm design \cite{hu2024flare, khetarpal2022toward, luo2024serl, rajeswaran2017learning, uchendu2023jump}. Recently, several works \cite{guo2025improving, lu2025vla} have demonstrated promising results by using RL to fine-tune pre-trained models. To optimize our MVEP (Multi-Viewpoint Exploration Policy), we designed pseudo-environment interactions and a view-aware reward function, enabling policy learning based on a pre-trained model using offline expert data. Furthermore, to adapt between the MVEP and the action policy, we introduce supervised post-training to coordinate the robot's ability to ``See and Act", thereby enhancing its task execution performance.

% \subsection{Active Vision-Action Model}

\section{Methodology}
\label{sec:method}

\subsection{Overview}

Our proposed \textbf{\ModelNameUpper\ (\ModelAbbr)} framework aims to identify better viewpoints for accurate and robust robotic manipulation, guided by task-specific visual feature extraction. The full pipeline is illustrated in Fig.~\ref{fig:framework}. \ModelAbbr\ takes as input a language instruction, the current visual observations from RGB-D cameras, and the current gripper state. To enable the model to explore the better observations from arbitrary viewpoints, we first reconstruct a 3D point cloud of the scene using the input RGB-D images. To narrow down the search space and enable task-aligned view selection, we leverage the coarse prediction stage from RVT-2~\cite{goyal2024rvt2} to generate an area of interest. However, unlike RVT-2, which extracts visual features for all tasks using a shared Multi-View Transformer (MVT), we integrate a \textbf{Task-Aware Mixture-of-Experts (TaskMoE)} module within the MVT to route instructions to specialized expert encoders. This design enables more precise and task-aligned visual feature extraction, which benefits both viewpoint selection and action prediction. Starting from the identified area of interest, we employ a \textbf{Multi-Viewpoint Exploration Policy (MVEP)} to search for better camera poses that maximize the visibility of the target object and the end-effector. The selected viewpoints are then re-rendered into image observations and processed using another TaskMoE-based MVT before being passed to the action prediction model. For action prediction, we also upgrade the autoregressive action policy proposed in ARP~\cite{zhang2025autoregressive} with our proposed TaskMoE to attain more task-specific feature extraction and action prediction.

In the following sections, we detail the design of:  
1) the TaskMoE,  
2) the MVEP,
and 3) the training strategy.

\subsection{TaskMoE}
\label{subsec:taskmoe}
\begin{figure}[t]
\centering
\includegraphics[width=0.9\columnwidth]{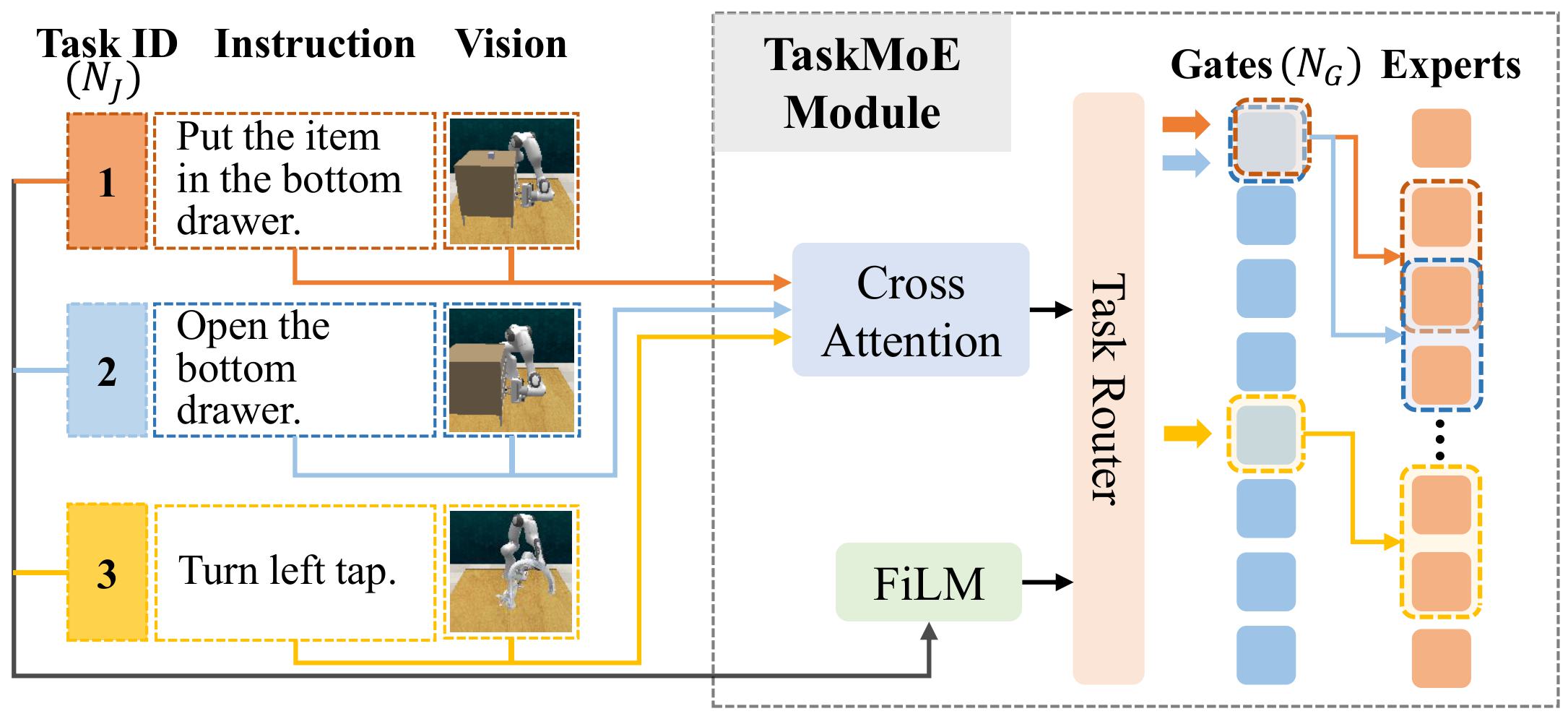}
\vspace{-10pt}
\caption{\textbf{Pipeline of the TaskMoE.} Our proposed TaskMoE takes \textit{Task ID}, \textit{Instruction}, and \textit{Vision} as inputs to guide expert selection for task-specific visual representation learning. To improve scalability and generalization, we design a compact gating mechanism with $N_G$ gates shared across $N_J$ tasks ($N_G < N_J$). This design allows tasks with similar action patterns (\eg, Task 1 and Task 2) to share the same gate, while assigning distinct gates to semantically diverse tasks (\eg, Task 3), thereby enabling effective feature specialization across a variety of manipulation tasks.}

\vspace{-15pt}
\label{fig:taskmoe}
\end{figure}

To address the inherent heterogeneity of complex manipulation tasks in multi-task learning, where different tasks often require substantially distinct visual representations and action policies, we introduce a task-aware Mixture-of-Experts module (TaskMoE), as illustrated in Fig.~\ref{fig:taskmoe}. Compared to prior MoE-based approaches~\cite{shazeer2017outrageously, wangsparse}, our TaskMoE introduces two key innovations.

First, instead of relying solely on task identifiers for expert selection, we incorporate richer instruction- and scene-related cues to guide expert routing more effectively, which is crucial for accurate multi-task robotic manipulation. Specifically, as shown in Fig.~\ref{fig:taskmoe}, we design a cross-modality module that employs a cross-attention mechanism to model the interaction between instruction and visual information. The resulting context-aware features are then fused with the task identifier via a Feature-wise Linear Modulation (FiLM) layer~\cite{Film}, enabling more adaptive and task-sensitive expert selection.

Second, to improve the scalability and generalization of TaskMoE, we decouple the number of routing gates from the total number of tasks. Concretely, we allocate $N_G$ gates for all $N_J$ tasks, where $N_G < N_J$. This design not only accommodates task diversity but also promotes parameter sharing among tasks with similar visual or semantic characteristics. For example, as illustrated in Fig.~\ref{fig:taskmoe}, \textit{Task 1} and \textit{Task 2} (both involving opening drawer) are routed through the same gate but directed to different experts based on their specific operation requirements. In contrast, \textit{Task 3}, which is semantically dissimilar, is routed through a different gate. This setup encourages the discovery of latent task clusters and provides the capacity to generalize to unseen tasks that share structural similarities with seen ones, thereby enhancing the transferability and robustness of TaskMoE. Notably, all tasks share a common pool of $N_E$ experts, and for each input, only the top-$k$ experts (based on gating scores) are activated to guide task-specific visual feature extraction.

\subsection{Multi-Viewpoint Exploration Policy}

Our proposed \textbf{Multi-Viewpoint Exploration Policy (MVEP)} aims to select $K$ viewpoints that maximally capture informative regions related to the manipulation target, thereby enhancing the accuracy of robotic action prediction. MVEPN takes as input the reconstructed point cloud $\mathcal{P} \in \mathbb{R}^{N \times 3}$ and its associated RGB features $\mathbf{F}_{\text{img}} \in \mathbb{R}^{N \times 3}$, which are concatenated to form the input representation:
\begin{equation}
\mathbf{X} = \text{Concat}(\mathcal{P}, \mathbf{F}_{\text{img}}) \in \mathbb{R}^{N \times 6},
\end{equation}
where $N$ denotes the number of 3D points. The fused representation $\mathbf{X}$ is processed by a multi-layer perceptron (MLP), which predicts the parameters of $K$ camera poses.

In this work, we represent each camera pose using a \textbf{look-at model}~\cite{lookatmodel}, which decouples the camera position and orientation via spherical coordinates. Therefore, the explored views are parameterized as 5-dimensional vectors $\mathbf{p}^{i}  = (\theta^{i}, \phi^{i}, r^{i}, \theta_\text{up}^{i}, \phi_\text{up}^{i}) \in \mathbb{R}^5$, where $(\theta, \phi, r)$ specify the camera center in spherical coordinates relative to the origin, $(\theta_\text{up}, \phi_\text{up})$ define the orientation of the up vector, and $i \in [0, K-1]$ means the $i_{th}$ view. So, all camera poses are $\mathbf{p} = \{\mathbf{p}^{i} | i=0, ..., K-1\}$. Additionally, the camera is perpetually oriented toward the coordinate origin $O(0, 0, 0)$.

% To facilitate gradient-based policy optimization using reinforcement learning, we model the camera pose parameters as samples from a Gaussian distribution. Instead of directly predicting the viewpoint $\mathbf{p}^i$, the network outputs the mean and log-standard deviation, enabling differentiable sampling via the reparameterization trick.

To facilitate gradient-based optimization of the viewpoint selection policy, we model each camera pose as a sample from a Gaussian distribution rather than predicting deterministic values. Specifically, for each of the $K$ viewpoints, our MVEPN outputs the mean and log-standard deviation of a diagonal Gaussian distribution over the 5-dimensional camera pose parameters. Formally, the output for the $i$-th viewpoint is denoted as $[\mu^i, \log \sigma^i]$, where:
\begin{equation}
    \begin{aligned}
        \mu^i &= [\mu_{\theta}^i, \mu_{\phi}^i, \mu_r^i, \mu_{\theta_{\text{up}}}^i, \mu_{\phi_{\text{up}}}^i], \\
        \log \sigma^i &= [\log \sigma_{\theta}^i, \log \sigma_{\phi}^i, \log \sigma_r^i, \log \sigma_{\theta_{\text{up}}}^i, \log \sigma_{\phi_{\text{up}}}^i].
    \end{aligned}
\end{equation}
Then the explored camera pose $\tilde{\mathbf{p}}^i = (\tilde{\theta}^i, \tilde{\phi}^i, \tilde{r}^i, \tilde{\theta}_{\text{up}}^i, \tilde{\phi}_{\text{up}}^i)$ is sampled using the reparameterization trick:
\begin{equation}
\tilde{\mathbf{p}}^i = \mu^i + \sigma^i \odot \epsilon^i, \quad \epsilon^i \sim \mathcal{N}(\mathbf{0}, \mathbf{I}),
\end{equation}
which enables end-to-end training via backpropagation.

To ensure the sampled poses remain valid in the spherical coordinate system, we constrain each component to lie within a normalized range using the sigmoid function:
\begin{equation}
    \begin{aligned}
        \tilde{\theta}^{i} &= \pi \cdot \sigma(\tilde{\theta}^{i}), \quad 
        \tilde{\phi}^{i} = 2\pi \cdot \sigma(\tilde{\phi}^{i}), \\
        \tilde{r}^{i} &= r_{\min} + (r_{\max} - r_{\min}) \cdot \sigma(\tilde{r}^{i}), \\
        \tilde{\theta}_{\text{up}}^{i} &= \pi \cdot \sigma(\tilde{\theta}_{\text{up}}^{i}), \quad 
        \tilde{\phi}_{\text{up}}^{i} = 2\pi \cdot \sigma(\tilde{\phi}_{\text{up}}^{i}),
    \end{aligned}
\end{equation}
where $\sigma(\cdot)$ denotes the sigmoid activation, and $r_{\min}$ and $r_{\max}$ define the allowable radial bounds.

\subsection{Training Strategy}

The entire training process of our \ModelNameUpper\ (\ModelAbbr) model consists of three stages: 

\noindent\textbf{Stage 1:} In the first stage, we train a fixed-view variant of \ModelAbbr\ using three default rendering perspectives: front, left, and top. This stage follows a similar training pipeline to RVT-2~\cite{goyal2024rvt2}, where the overall loss comprises several components. Specifically, $\mathcal{L}_{hc}$ and $\mathcal{L}_{hf}$ denote the cross-entropy losses over heatmaps produced by the coarse and fine grounding modules, respectively. The ground-truth heatmaps are generated using a truncated Gaussian distribution centered at the 2D projection of the ground-truth 3D location, as described in RVT~\cite{goyal2023rvt}. Additionally, the loss includes the end-effector rotation loss $\mathcal{L}_{rot}$, computed as the cross-entropy loss for each of the Euler angles, as well as the gripper state loss $\mathcal{L}_{gri}$ and collision indicator loss $\mathcal{L}_{col}$, both formulated as binary classification losses. The total objective for the first stage is given by:
\begin{equation}
    \mathcal{L}_{\text{s1}} = \mathcal{L}_{hc} + \mathcal{L}_{hf} + \mathcal{L}_{rot} + \mathcal{L}_{gri} +  \mathcal{L}_{col}.
    \label{eq:l_s1}
\end{equation}

\noindent\textbf{Stage 2:} In the second stage, we refine the Multi-View Exploration Policy (MVEP), using the fixed-view \ModelAbbr\ model trained in Stage 1 as a baseline. MVEP is optimized via the Proximal Policy Optimization (PPO) algorithm \cite{schulman2017proximal}, which typically requires interaction with a physical environment to obtain reward feedback. To mitigate the high time cost, we introduce a novel \textit{pseudo-environment interaction} mechanism. The details of the mechanism are presented in the supplementary materials.

Specifically, we treat the fixed-view \ModelAbbr\ as a reference model to supervise the training of MVEP. Given the same observations and instructions, the reference model yields losses $\mathcal{L}_{\text{ref}} = [\mathcal{L}_{\text{hf}}, \mathcal{L}_{\text{rot}}, \mathcal{L}_{\text{gri}}, \mathcal{L}_{\text{col}}]$. In contrast, MVEP generates dynamic viewpoints for action prediction, resulting in losses $\mathcal{L}_{\text{\ModelAbbr}} = [\mathcal{L}'_{\text{hf}}, \mathcal{L}'_{\text{rot}}, \mathcal{L}'_{\text{gri}}, \mathcal{L}'_{\text{col}}]$. Taking the reference model as a performance lower bound, our objective is to make $\mathcal{L}_{\text{\ModelAbbr}}$ smaller than $\mathcal{L}_{\text{ref}}$. The first reward term is therefore defined as:
\begin{equation}
    r_{0} = \mathcal{L}_{\text{ref}} - \mathcal{L}_{\text{\ModelAbbr}}.
\end{equation}
In addition to the task loss reward, we incorporate a confidence-based reward derived from the fine grounding module. Specifically, we compute the negative average entropy across the grounding heatmaps:
\begin{equation}
    r_{1} = -\frac{1}{K}\sum_{i=1}^{K} \mathcal{H}(\text{softmax}(\mathbf{H}_i)),
\end{equation}
where $\mathcal{H}$ denotes the entropy function and $\mathbf{H}_i$ is the heatmap for view $i$ from the fine grounding module.

To further encourage viewpoint diversity, we introduce an additional reward based on the average pairwise cosine distance between camera positions:
\begin{equation}
    r_{2} = \frac{1}{K(K-1)}\sum_{i \neq j} \left(1 - \cos(\mathbf{p}_i, \mathbf{p}_j)\right),
\end{equation}
where $\mathbf{p}_i$ represents the camera position for view $i$.

The overall reward is computed by adaptively normalizing and aggregating all components:
\begin{equation}
    r = \sum_{i=0}^{2} w_i \cdot \mathcal{N}(r_i),
\end{equation}
where $w_i$ are learnable component weights and $\mathcal{N}$ is an online normalization operator. The adaptive normalization maintains the mean and variance of each component using the Welford algorithm \cite{welford1962note} for normalization, and clips the reward to the range $[-10, 10]$ to ensure training stability.

In this stage, only MVEP is trainable, and all the other components of \ModelAbbr\ are frozen.

\noindent\textbf{Stage 3:} To enable MVEP to better adapt to action generation, we further fine-tune the entire \ModelAbbr\ model, excluding MVEP, using the same loss functions as in Stage 1.

\begin{table*}[t]
\centering
\scriptsize
\resizebox{1.005\linewidth}{!}{%
\small
\setlength{\tabcolsep}{2.0pt}
\begin{tabular}{lcccccccccc}
\toprule
% Method & \multicolumn{2}{c}{\textbf{Avg. SR (\%)} $\uparrow$} & Close Jar & Drag Stick & Insert Peg & Meat off Grill & Open Drawer & Place Cups & Place Wine & Push Buttons \\  
Method & \textbf{Avg. SR (\%)} $\uparrow$ & \textbf{Avg. Rank} $\downarrow$ & Close Jar & Drag Stick & Insert Peg & Meat off Grill & Open Drawer & Place Cups & Place Wine & Push Buttons \\
\midrule
C2F-ARM-BC \cite{james2022coarse} (CVPR) & 20.1 & 9.67 & 24.0 & 24.0 & 4.0 & 20.0 & 20.0 & 0.0 & 8.0 & 72.0 \\
PerAct \cite{shridhar2023perceiver} (CoRL) & 49.4 & 7.06 & $55.2{\pm}4.7$ & $89.6{\pm}4.1$ & $5.6{\pm}4.1$ & $70.4{\pm}2.0$ & $88.0{\pm}5.7$ & $2.4{\pm}3.2$ & $44.8{\pm}7.8$ & $92.8{\pm}3.0$ \\
HiveFormer \cite{guhur2023instruction} (CoRL) & 45.0 & 8.22 & 52.0 & 76.0 & 0.0 & 80.0 & 52.0 & 0.0 & 80.0 & 84.0 \\
PolarNet \cite{chen2023polarnet} (CoRL) & 46.0 & 7.06 &  36.0 & 92.0 & 4.0 & \textbf{100.0} & 84.0 & 0.0 & 40.0 & 96.0 \\
RVT \cite{goyal2023rvt} (CoRL) & 62.9 &  5.28 & $52.0{\pm}2.5$ & $99.2{\pm}1.6$ & $11.2{\pm}3.0$ & $88.0{\pm}2.5$ & $71.2{\pm}6.9$ & $4.0{\pm}2.5$ & $91.0{\pm}5.2$ & $\mathbf{100.0{\pm}0.0}$ \\
Act3D \cite{gervet2023act3d} (CoRL) & 63.2 & 5.56 & $96.8{\pm}3.0$ & $80.8{\pm}6.4$ & $24.0{\pm}8.4$ & $95.2{\pm}1.6$ & $78.4{\pm}11.2$ & $3.2{\pm}3.0$ & $59.2{\pm}9.3$ & $93.6{\pm}2.0$ \\
3D Diffuser Actor \cite{ke20253d} (CoRL) & 81.3 & 3.0 & $96.0{\pm}2.5$ & $\mathbf{100.0{\pm}0.0}$ & $65.6{\pm}4.1$ & $96.8{\pm}1.6$ & $89.6{\pm}4.1$ & $24.0{\pm}7.6$ & $93.6{\pm}4.8$ & $98.4{\pm}2.0$ \\
% \midrule
RVT2 \cite{goyal2024rvt2} (RSS) & 81.4 & 2.89 & $\mathbf{100.0{\pm}0.0}$ & $99.0{\pm}1.7$ & $40.0{\pm}0.0$ & $\mathbf{99.0{\pm}1.0}$ & $74.0{\pm}11.8$ & $38.0{\pm}4.5$ & $\mathbf{95.0{\pm}3.3}$ & $\mathbf{100.0{\pm}0.0}$ \\
ARP \cite{zhang2025autoregressive} (RA-L) & 81.6 & 2.83 & 97.6 & 88.0 & 53.2 & $96.0$ & $\mathbf{90.4}$ & $48.0$ & $92.0$ & $\mathbf{100.0}$ \\
% TVVE$^{--}$ (w/o MVEP \& Rendering) & \multicolumn{2}{c}{--} & -- & -- & -- & -- & -- & -- & -- & --\\
\textbf{\ModelAbbr\ (Ours)} & \textbf{86.6} & \textbf{2.17} & $\mathbf{100.0{\pm}0.0}$ & $\mathbf{100.0{\pm}0.0}$ & $\mathbf{98.0{\pm}2.8}$ & $94.0{\pm}2.8$ & $90.0{\pm}2.8$ & $\mathbf{54.0{\pm}2.8}$ & $92.0{\pm}5.7$ & $\mathbf{100.0{\pm}0.0}$ \\
\midrule
Method & Put in Cupboard & Put in Drawer & Put in Safe & Screw Bulb & Slide Block & Sort Shape & Stack Blocks & Stack Cups & Sweep to Dustpan & Turn Tap \\
\midrule
C2F-ARM-BC \cite{james2022coarse} (CVPR) & 0.0 & 4.0 & 12.0 & 8.0 & 16.0 & 8.0 & 0.0 & 0.0 & 0.0 & 68.0 \\
PerAct \cite{shridhar2023perceiver} (CoRL) & $28.0{\pm}4.4$ & $51.2{\pm}4.7$ & $84.0{\pm}3.6$ & $17.6{\pm}2.0$ & $74.0{\pm}13.0$ & $16.8{\pm}4.7$ & $26.4{\pm}3.2$ & $2.4{\pm}2.0$ & $52.0{\pm}0.0$ & $88.0{\pm}4.4$ \\
HiveFormer \cite{guhur2023instruction} (CoRL) & 32.0 & 68.0 & 76.0 & 8.0 & 64.0 & 12.0 & 4.0 & 0.0 & 28.0 & 80.0 \\
PolarNet \cite{chen2023polarnet} (CoRL) & 12.0 & 32.0 & 84.0 & 44.0 & 56.0 & 12.0 & 8.0 & 8.0 & 52.0 & 80.0 \\
RVT \cite{goyal2023rvt} (CoRL) & $49.6{\pm}3.2$ & $88.0{\pm}5.7$ & $91.2{\pm}3.0$ & $48.0{\pm}5.7$ & $81.6{\pm}5.4$ & $36.0{\pm}2.5$ & $28.8{\pm}3.9$ & $26.4{\pm}8.2$ & $72.0{\pm}0.0$ & $93.6{\pm}4.1$ \\
Act3D \cite{gervet2023act3d} (CoRL) & $67.2{\pm}3.0$ & $91.2{\pm}6.9$ & $95.2{\pm}4.0$ & $32.8{\pm}6.9$ & $96.0{\pm}2.5$ & $29.6{\pm}3.2$ & $4.0{\pm}3.6$ & $9.6{\pm}6.0$ & $86.4{\pm}6.5$ & $94.4{\pm}2.0$ \\
3D Diffuser Actor \cite{ke20253d} (CoRL) & $\mathbf{85.6{\pm}4.1}$ & $96.0{\pm}3.6$ & $\mathbf{97.6{\pm}2.0}$ & $82.4{\pm}2.0$ & $97.6{\pm}3.2$ & $44.0{\pm}4.4$ & $68.3{\pm}3.3$ & $47.2{\pm}8.5$ & $84.0{\pm}4.4$ & $99.2{\pm}1.6$ \\
% \midrule
RVT2 \cite{goyal2024rvt2} (RSS) & $66.0{\pm}4.5$ & $96.0{\pm}0.0$ & $96.0{\pm}2.8$ & $\mathbf{88.0{\pm}4.9}$ & $92.0{\pm}2.8$ & $35.0{\pm}2.8$ & $\mathbf{80.0{\pm}2.8}$ & $\mathbf{69.0{\pm}5.9}$ & $\mathbf{100.0{\pm}0.0}$ & $99.0{\pm}1.7$ \\
ARP \cite{zhang2025autoregressive} (RA-L) & 68.0 & $99.2$ & 94.4 & $85.6$ & $98.4$ & $35.2$ & 55.2 & $76.8$ & $90.4$ & $\mathbf{100.0}$ \\
% TVVE$^{--}$ (w/o MVEP \& Rendering)  & -- & -- & -- & -- & -- & -- & -- & -- & -- & -- \\
\textbf{\ModelAbbr\ (Ours)} & $74.0{\pm}8.5$ & $\mathbf{100.0{\pm}0.0}$ & $78.0{\pm}2.8$ & $86.0{\pm}2.8$ & $\mathbf{100.0{\pm}0.0}$ & $\mathbf{62.0{\pm}8.5}$ & $74.0{\pm}2.8$ & $64.0{\pm}5.7$ & $92.0{\pm}5.7$ & $\mathbf{100.0{\pm}0.0}$ \\
\bottomrule
\end{tabular}%
}
\vspace{-10pt}
\caption{\textbf{Performance comparison across 18 tasks in RLBench under the multi-view setup.} Our method achieves a substantial improvement in the average success rate. Moreover, the lower average rank of 2.17 demonstrates the superior overall performance and better generalization ability of \ModelAbbr\ across all 18 tasks.}

\vspace{-8pt}
\label{table:rlb-performance}
\end{table*}

\begin{table*}[h]
\centering
\scriptsize
\resizebox{1.005\linewidth}{!}{%
\small
\setlength{\tabcolsep}{2.0pt}
\begin{tabular}{lccccccccccccc}
\toprule
& \textbf{Avg. SR (\%)} $\uparrow$ & \textbf{Avg. Rank} $\downarrow$ & close jar & open drawer & sweep to dustpan & meat off grill & turn tap & slide block & put in drawer & drag stick & push buttons & stack blocks \\
\midrule
GNFactor~\cite{ze2023gnfactor} (CoRL) & 31.7 & 4.90 & 25.3 & 76.0 & 28.0 & 57.3 & 50.7 & 20.0 & 0.0 & 37.3 & 18.7 & 4.0 \\
Act3D~\cite{gervet2023act3d} (CoRL) & 65.3 & 3.35 & $52.0_{\pm5.7}$ & $84.0_{\pm8.6}$ & $80.0_{\pm9.8}$ & $66.7_{\pm1.9}$ & $64.0_{\pm5.7}$ & $\mathbf{100.0_{\pm0.0}}$ & $54.7_{\pm3.8}$ & $86.7_{\pm1.9}$ & $64.0_{\pm1.9}$ & $0.0_{\pm0.0}$ \\
3D Diffuser Actor~\cite{ke20253d} (CoRL) & 78.4 & 2.20 & 82.7$_{\pm1.9}$ & $\mathbf{89.3_{\pm7.5}}$ & $\mathbf{94.7_{\pm1.9}}$ & 88.0$_{\pm5.7}$ & 80.0$_{\pm8.6}$ & 92.0$_{\pm0.0}$ & 77.3$_{\pm3.8}$ & $\mathbf{98.7_{\pm1.9}}$ & 69.3$_{\pm5.0}$ & 12.0$_{\pm3.7}$ \\
RVT2 \cite{goyal2024rvt2} (RSS)  & $82.7_{\pm0.7}$ & 2.40 & $96.0_{\pm0.0}$ & $78.7_{\pm3.8}$ & $73.3_{\pm5.0}$ & $\mathbf{93.3_{\pm1.9}}$ & $\mathbf{93.3_{\pm3.8}}$ & $89.3_{\pm1.9}$ & $73.3_{\pm1.9}$ & $76.0_{\pm0.0}$ & $\mathbf{96.0_{\pm0.0}}$ & $\mathbf{57.3_{\pm3.8}}$ \\
\textbf{\ModelAbbr\ (Ours)}  & $\mathbf{83.2_{\pm1.7}}$ & \textbf{2.15} & $\mathbf{100.0_{\pm0.0}}$ & $84.0_{\pm5.7}$ & $94.0_{\pm2.8}$ & $92.0_{\pm0.0}$ & $92.0_{\pm5.7}$ & $90.0_{\pm2.8}$ & $\mathbf{98.0_{\pm2.8}}$ & $50.0_{\pm2.8}$ & $82.0_{\pm2.8}$ & $50.0_{\pm8.5}$ \\
\bottomrule
\end{tabular}
}
\vspace{-8pt}
\caption{\textbf{Performance comparison across 10 tasks in RLBench under the single-view setup.} Our~\ModelAbbr\ achieves a substantial improvement in the average success rate compared to previous methods.}
\vspace{-15pt}
\label{table:rlb-performance-single-view}
\end{table*}

\section{Experiments}
\label{{sec:exp}}

To comprehensively assess the effectiveness and generalization of our \ModelNameUpper (\ModelAbbr), we perform extensive experiments across both simulated and real-world environments.

% Detailed experimental settings for the real-world and the corresponding results are presented in the supplementary materials.

\subsection{Datasets and Experimental Setups}
\label{sec:exp_setup}
% \subsubsection{Simulation Setup.}
\noindent\textbf{Simulation.}
% \label{sec:simulation-setup}
Our experiments are conducted on \textbf{RLBench} dataset~\cite{james2020rlbench} and \textbf{RLBench-OG}. 

\textbf{RLBench} is a multi-task learning benchmark implemented in the CoppeliaSim simulator featuring a 7-degree-of-freedom (7-DoF) Franka Emika Panda robot in a tabletop environment. Its observations consist of RGB-D images ($128 \times 128$) captured from four fixed views, including frontal, left shoulder, right shoulder, and wrist-mounted. The action space represents the end-effector pose as $\mathbf{a} = (\mathbf{p}_{\text{xyz}}, \mathbf{q}_{\text{rot}}, g_{\text{gripper}}, c_{\text{collision}}) \in \mathbb{R}^7 \times \{0,1\}^2$, where $\mathbf{p}_{\text{xyz}} \in \mathbb{R}^3$ denotes Cartesian coordinates, $\mathbf{q}_{\text{rot}} \in \mathbb{R}^4$ is the quaternion rotation, $g_{\text{gripper}} \in \{0,1\}$ controls gripper state, and $c_{\text{collision}} \in \{0,1\}$ indicates collision. 
Following previous works~\cite{goyal2024rvt2,ze2023gnfactor}, we evaluate our method under two standard configurations: (1) a \textit{multi-view setup} consisting of 18 manipulation tasks (2–60 variations each) with all four camera views, and (2) a \textit{single-view setup} including 10 tasks observed only from the front RGB-D camera. The specific tasks are provided in the supplementary materials.

% For both setups, we report the mean and standard deviation of the success rates over 25 episodes per task, and all tests are repeated three times. 

% \textbf{RLBench-OG} is derived from RLBench and is used to evaluate the robustness and generalization of our \ModelAbbr\ under occluded scenarios and complex environments. It comprises two distinct suites: the \textit{Occlusion Suite} and the \textit{Generalization Suite}, with variants created across 10 manipulation tasks during evaluation. 

\textbf{RLBench-OG} is derived from RLBench and used for validate the robustness of our \ModelAbbr\ under occluded scenarios and its generalization capability in complex environments. RLBench-OG consists of two distinct suites: the \textit{Occlusion Suite} and the \textit{Generalization Suite}. We created variants across 10 tasks for evaluation. In the Occlusion Suite, we designed two experimental configurations: 1) Models are both trained and tested directly under the occluded task configurations; 2) Models are trained in the original RLBench task settings and then evaluated in a zero-shot manner under occluded conditions. For the Generalization Suite, models are trained in the original RLBench environment and subsequently evaluated in a zero-shot manner across various generalization settings. Each task is configured with 50 episodes for training. The mean success rate, evaluated on 25 episodes per task, is reported as mean $\pm$ standard deviation across three independent experimental runs. For simulating occlusion, we introduce occluders or rotate manipulated objects to create occlusions from a frontal viewpoint. For generalization, we introduce perturbations by: 1) altering scene lighting, 2) changing the color and texture of the tabletop, 3) modifying the color and texture of the background, 4) adding distractors, and 5) adjusting the pose of the observation camera. For further details on RLBench-OG, please refer to the supplementary material.

\noindent\textbf{Real World.}
To evaluate the generalization of our \ModelAbbr\ across different robotic platforms, we conducted experiments on both the \textit{Franka Research 3} and the \textit{Dobot Nova 2}. For the Franka Research 3, a single third-person camera (ORBBEC Femto Bolt) was mounted directly in front of the arm. For the Dobot Nova 2, three Intel RealSense depth cameras were used: an overhead lateral camera (D435i), a front lateral camera (D455), and a wrist-mounted camera on the right arm (D405), as illustrated in the supplementary materials.
We designed five distinct manipulation tasks: \textit{Pick Grape}, \textit{Stack Bowls}, \textit{Push Buttons}, \textit{Collect Fruits}, and \textit{Put Item In Drawer}. Expert demonstrations were collected with 50 episodes per task for training. To improve efficiency, only key frames from the demonstrations were retained. Additional details of the real-world experiments are provided in the supplementary materials.

% \subsubsection{Implementation Settings.}
% \noindent\textbf{Implementation Settings.}
% In our experiments, the number of camera viewpoints $K=3$, the number of gates in TaskMoE $N_{G} = 8$, and the number of task experts $N_{E} = 16$. For each task, the $Top-2$ task experts are selected. Our \ModelAbbr\ model was trained using $4 \times$ NVIDIA RTX A800 GPUs with 80GB memory. During testing, we sequentially validated each task using $1 \times$ NVIDIA RTX A800 GPU and computed the average success rate. We provide the mean and standard deviation of the success rates over 25 episodes per task, and all tests are repeated three times. 

\noindent\textbf{Implementation Settings.}
In our experiments, we set the number of camera viewpoints to $K = 3$, the number of gates in TaskMoE to $N_{G} = 8$, and the number of task experts to $N_{E} = 16$. For each task, the top-2 experts are selected. \ModelAbbr\ was trained on $4$ NVIDIA RTX A800 GPUs. During testing, tasks were sequentially evaluated on a single NVIDIA RTX A800 GPU, and the average success rate was computed.

% The default radial constraint for the camera is $r_{\text{min}} = 0.75$~m and $r_{\text{max}} = 1.3$~m. 
\begin{figure*}[!t]
    \centering
\includegraphics[width=0.95\linewidth]{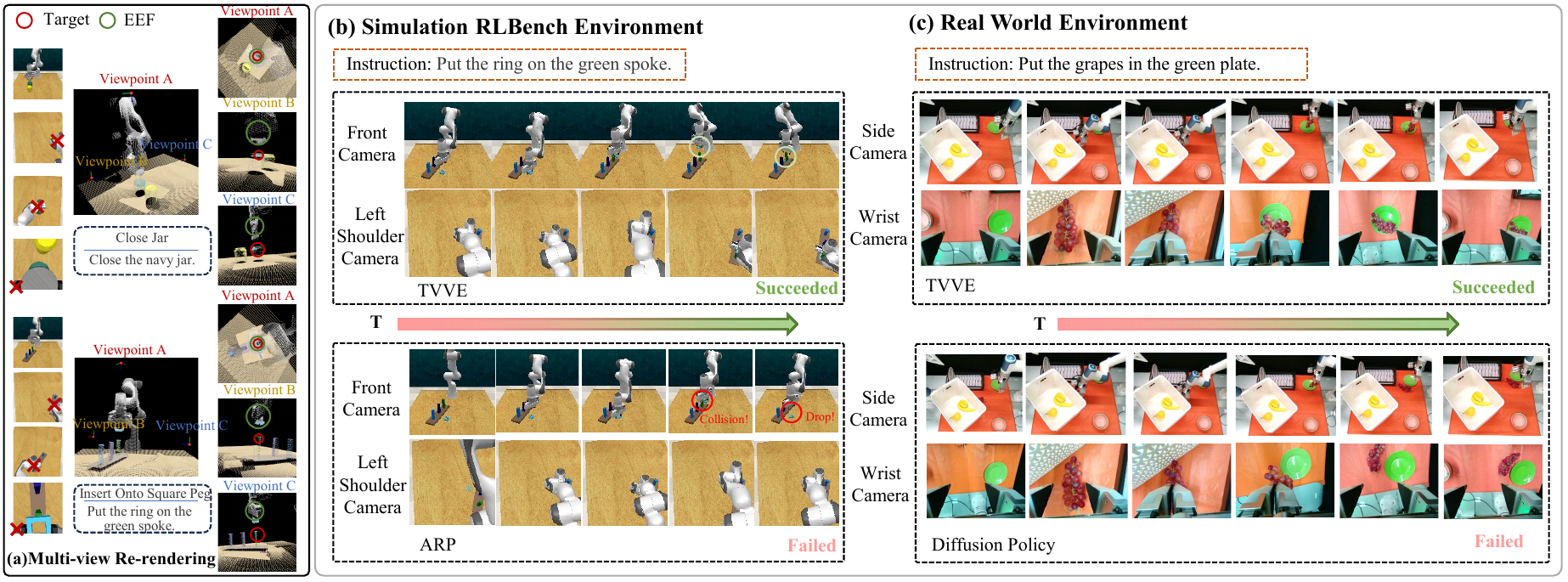}
\vspace{-10pt}
    \caption{Visualization of our \ModelAbbr\ in the simulation RLBench Environment and Diffusion Policy in the Real-world Environment. (a) In RLBench, we visualize the re-rendering results of dynamic multi-view in the scenes for the \textit{Close Jar} and \textit{Insert Peg} tasks, where EEF in the figures denotes the end-effector. Visualizations of additional tasks are provided in the supplementary material. (b) Sample of \textit{Inser peg}. (c) Sample of \textit{Pick Grape} in the real world. }
    \vspace{-8pt}
    \label{fig:vis}
\end{figure*}

\begin{table*}[!t]
\centering
\scriptsize
\resizebox{1.0\linewidth}{!}{%
\small
\setlength{\tabcolsep}{2.0pt}
\begin{tabular}{lcccccccccc}
\toprule
\textbf{Models} & \textbf{Avg. SR (\%)} $\uparrow$ & \textbf{Avg. Rank} $\downarrow$
& \textbf{Occlusion 1} & \textbf{Occlusion 2} & \textbf{Light Color} & \textbf{Table Color} &
\textbf{Table Texture} & \textbf{Distractor} & \textbf{Background Texture} & \textbf{Camera Pose} \\
\midrule
Diffusion Policy \cite{chi2023diffusion} (IJRR) &
$ 23.8\pm1.9 $ & 4.0 &
$ 27.4\pm0.8 $ & $ 23.4\pm1.1 $ & $ 22.9\pm0.8 $ & $ 22.5\pm0.7 $ &
$ 22.6\pm0.5 $ & $ 24.4\pm1.4 $ & $ 24.9\pm1.2 $ & $ 22.2\pm1.2 $ \\
ARP \cite{zhang2025autoregressive} (RA-L) &
$ 63.7\pm6.1 $ & 2.5 &
$ 73.0\pm1.6 $ & $ 52.6\pm1.0 $ & $ 59.8\pm0.8 $ & $ 62.7\pm0.5 $ &
$ 61.3\pm1.0 $ & $ 62.4\pm1.4 $ & $ 68.1\pm1.6 $ & $ 69.7\pm1.4 $ \\
RVT2 \cite{goyal2024rvt2} (RSS) &
$ 64.5\pm8.3 $ & 2.1 &
$ 72.8\pm2.0 $ & $ 46.9\pm0.4 $ & $ 60.8\pm1.2 $ & $ 61.8\pm1.1 $ &
$ 64.0\pm2.0 $ & $ \mathbf{63.4\pm0.2} $ & $ 72.6\pm0.6 $ & $ \mathbf{74.0\pm0.8} $ \\
\textbf{\ModelAbbr\ (Ours)} &
$\mathbf{67.0\pm6.2}$ & \textbf{1.4} &
$\mathbf{75.0\pm1.2}$ & $\mathbf{58.0\pm1.1}$ & $\mathbf{63.7\pm1.1}$ &
$\mathbf{64.6\pm0.7}$ & $\mathbf{66.8\pm0.9}$ & $ 60.2\pm0.6 $ &
$\mathbf{74.3\pm0.9}$ & $ 73.2\pm0.9 $ \\
\bottomrule
\end{tabular}
}
\vspace{-8pt}
\caption{Comprehensive evaluation of model performance under occlusion and generalization settings in \textbf{RLBench-OG}, covering both zero-shot and trained scenarios. Detailed per-task success rates are provided in the supplementary material.}
\vspace{-10pt}
\label{table:rlb-og-performance}
\end{table*}

% \subsection{Main Results}
\subsection{Comparison with State-of-the-art Methods}

To comprehensively evaluate the effectiveness of our ~\ModelAbbr, we conduct comparisons with a broad range of state-of-the-art approaches, including Diffusion Policy~\cite{chi2023diffusion}, C2F-ARM-BC~\cite{james2022coarse}, PerAct~\cite{shridhar2023perceiver}, Hiveformer~\cite{guhur2023instruction}, GNFactor~\cite{ze2023gnfactor}, RVT~\cite{goyal2023rvt}, RVT2~\cite{goyal2024rvt2}, Act3D~\cite{gervet2023act3d}, 3D Diffuser Actor~\cite{ke20253d}, and ARP~\cite{zhang2025autoregressive}.
We evaluate them both quantitatively and qualitatively across three benchmarks: RLBench, RLBench-OG, and real-world manipulation tasks.

% \textbf{Results Across 18 RLBench Tasks.}
\noindent\textbf{Comparisons on RLBench.}
Tables~\ref{table:rlb-performance} and~\ref{table:rlb-performance-single-view} present the quantitative results on RLBench~\cite{james2020rlbench} under the multi-view and single-view setups, respectively. Our proposed \ModelAbbr\ consistently achieves significant improvements in average success rate across both settings. Specifically, \ModelAbbr\ attains an average success rate of \textbf{86.6\%} under the multi-view setup, surpassing the previous state-of-the-art method ARP~\cite{zhang2025autoregressive} by approximately \textbf{5\%}. Notably, substantial gains are observed in tasks such as \textit{Insert Peg} (from $65.6\%$ to $98.0\%$) and \textit{Sort Shape} (from $36.0\%$ to $62.0\%$). A qualitative example of \textit{Insert Peg} is illustrated in Fig.~\ref{fig:vis}(a)(b). It shows that compared to ARP~\cite{zhang2025autoregressive}, which relies on fixed viewpoints, our \ModelAbbr\ dynamically renders more informative perspectives, effectively mitigating occlusions and collisions while providing richer context for more accurate action prediction, validating that dynamic ``seeing" underpins robust ``acting." Moreover, as indicated by the lower average rank of \textbf{2.17} in Table~\ref{table:rlb-performance}, \ModelAbbr\ demonstrates superior overall performance across all 18 tasks, highlighting its enhanced generalization capability, which is primarily attributed to the proposed TaskMoE design.

\begin{table}[t]
\centering
\scriptsize
\resizebox{1.005\linewidth}{!}{%
\small
\setlength{\tabcolsep}{2.0pt}
\begin{tabular}{llc|ccccc}
\toprule
\textbf{Arm} & \textbf{Method} & \textbf{Avg. SR (\%)} $\uparrow$ & \textbf{Pick Grape} & \textbf{Stack Bowls} & \textbf{Push Buttons} & \textbf{Collect Fruits} & \textbf{Put Item In Drawer} \\
\midrule
\multirow{2}{*}{Dobot} 
& Diffusion Policy & 68.0 & 90.0 & 70.0 & 70.0 & 50.0 & 60.0 \\
& \textbf{\ModelAbbr\ (Ours)} & \textbf{88.0} & \textbf{100.0} & \textbf{90.0} & \textbf{100.0} & \textbf{70.0} & \textbf{80.0} \\
\midrule
\multirow{3}{*}{Franka} 
& Diffusion Policy & 58.0 & 90.0 & 50.0 & 60.0 & 40.0 & 50.0 \\
& ARP & 72.0 & 100.0 & 70.0 & 80.0 & 50.0 & 60.0 \\
& \textbf{\ModelAbbr\ (Ours)} & \textbf{78.0} & \textbf{100.0} & \textbf{70.0} & \textbf{90.0} & \textbf{60.0} & \textbf{70.0} \\
\bottomrule
\end{tabular}
}
\vspace{-8pt}
\caption{Comparison of task success rates for Dobot and Franka robots across different manipulation tasks. For more real-robot experiments, including robustness and generalization testing, please refer to the supplementary materials. \textsuperscript{*}Franka and Dobot adopt different experimental configurations. }
\vspace{-15pt}
\label{tab:task_comparison_in_real_world}
\end{table}

\noindent\textbf{Comparisons on RLBench-OG.} Table~\ref{table:rlb-og-performance} highlights that \ModelAbbr\ achieves the best overall success rate (67.0\% $\pm$ 6.2) and the best average rank (1.1) on RLBench-OG, surpassing ARP (63.7\% $\pm$ 6.1, rank 1.8) and the Diffusion Policy baseline (23.8\% $\pm$ 1.9, rank 3.0). The advantage of our virtual-view exploration, dynamic virtual multi-view re-rendering and TaskMoE manifests consistently across perturbations: \ModelAbbr\ yields +2.0/+5.4 absolute gains over ARP for Occlusion~1/2, maintains higher resilience to background texture shifts (74.3\% $\pm$ 0.9 vs.\ 68.1\% $\pm$ 1.6), and preserves reliable execution under camera pose changes (73.2\% $\pm$ 0.9 vs.\ 69.7\% $\pm$ 1.4). These improvements confirm that learning task-aware viewpoints and decoupled visual representations mitigates the interference issues, leading to more robust manipulation under severe OOD disturbances.

\noindent\textbf{Comparisons on Real-World Tasks.}
In Dobot robot experiment, compared to the baseline method Diffusion Policy (DP) \cite{chi2023diffusion}, \ModelAbbr\ achieves superior performance across all tasks, attaining a significantly higher average success rate of 88.0\% versus 68.0\% for DP. Notable gains are observed in tasks such as ``Push Buttons" (+30\%) and ``Stack Bowls" (+20\%), underscoring the effectiveness of integrating task-aware visual planning into the policy framework. A qualitative example of \textit{Pick Grape} is illustrated in Fig.~\ref{fig:vis}(c). In Franka robot experiment, \ModelAbbr\ achieves the highest average success rate of 78.0\%, outperforming both Diffusion Policy (58.0\%) and ARP (72.0\%). These results further demonstrate its strong adaptability in real-world scenarios. We report the success rates based on 10 trials for each task, with detailed results provided in Table \ref{tab:task_comparison_in_real_world}. More real-world results are shown in the supplementary material.

\begin{table}[t]
\centering\scriptsize
\setlength{\tabcolsep}{20pt}
\begin{tabular}{l c}
\toprule
\textbf{Configuration} & \textbf{Avg. SR (\%)} $\uparrow$ \\
\midrule
\textbf{\ModelAbbr}~(TaskMoE$+$MVEP) & \textbf{86.6} \\
w/o TaskMoE & 85.6 \\
\makecell[l]{Fixed Viewpoints} & 83.3 \\
\makecell[l]{Random Viewpoints} & 8.9 \\
\midrule
\end{tabular}
\vspace{-8pt}
\caption{\textbf{Ablation study results on RLBench \textit{multi-view setup}.} Removing TaskMoE or replacing the explored viewpoints attained by MVEP with random or fixed ones degrades manipulation performance compared to the full \ModelAbbr\ with both components.}
\vspace{-8pt}
\label{tab:ablation}
\end{table}

\subsubsection{Ablation Studies.}
In this section, we further analyze the impact of the TaskMoE, Multi-Viewpoint Exploration Policy (MVEP), and the hyperparameters $K$ and $r$, supported by quantitative results in Tables~\ref{tab:ablation},~\ref{tab:womoe}, and~\ref{tab:camera_constraints}. Specifically, the results in Table~\ref{tab:ablation} are obtained from experiments on all 18 RLBench tasks, while those in Tables~\ref{tab:womoe} and~\ref{tab:camera_constraints} are based on five representative tasks selected from RLBench, namely \textit{Put Item in Drawer}, \textit{Turn Tap}, \textit{Put Groceries in Cupboard}, \textit{Put Money in Safe}, and \textit{Close Jar}.

\noindent\textbf{Effectiveness of TaskMoE.}
The TaskMoE module is designed to enhance multi-task generalization. From Table~\ref{tab:ablation}, removing TaskMoE from both the feature extraction and action generation components (\textit{w/o TaskMoE}) leads to a moderate performance drop to 85.6\%, underscoring its importance for multi-task representation learning. 
To further assess its generalization ability to OOD tasks unseen during training, additional results are presented in Table~\ref{tab:womoe}. The results show that TaskMoE not only improves performance on trained tasks but also facilitates transfer to unseen ones. Specifically, on tasks encountered during training, the model equipped with TaskMoE achieves an average success rate of 80.8\%, outperforming the variant without TaskMoE (72.0\%). This demonstrates its ability to learn and execute diverse, known tasks more effectively. For unseen tasks, the model without TaskMoE achieves a success rate of 60.0\% on the new task \textit{Open Drawer}, while the one with TaskMoE attains a higher success rate of 72.0\%, highlighting its enhanced generalization capability.

\noindent\textbf{Effectiveness of MVEP.}
The Multi-Viewpoint Exploration Policy (MVEP) is designed to dynamically explore better camera viewpoints that reduce occlusion and provide more informative visual feature for improved robotic manipulation. To evaluate the effectiveness of MVEP, we replace the explored viewpoints with either fixed viewpoints used in RVT2~\cite{goyal2024rvt2} or randomly selected ones. As shown in Table~\ref{tab:ablation}, the manipulation success rate drops from $86.6\%$ to $83.3\%$ when using fixed viewpoints, and further plummets to $8.9\%$ when using random viewpoints. This significant degradation demonstrates that the learned viewpoint exploration policy in MVEP effectively identifies informative perspectives, which are crucial for accurate perception and robust action execution.

\begin{table}[t]
\centering
\scriptsize
\setlength{\tabcolsep}{4pt}
\begin{tabular}{l|cccccc|cc}
\toprule
% \rowcolor{lightgray}
& \multicolumn{6}{c|}{\textbf{Seen}} & \textbf{Unseen} \\
\textbf{TaskMoE}
& \textbf{Avg. SR (\%)} $\uparrow$ & \textbf{PID} & \textbf{TT} & \textbf{PGC} & \textbf{PMS} & \textbf{CJ} 
& \begin{tabular}[c]{@{}c@{}}Open drawer\end{tabular} \\
\midrule
\checkmark & 80.8 & 100.0 & 100.0 & 32.0 & 88.0 & 84.0 & 72.0 \\
$\times$  & 72.0 & 96.0 & 96.0 & 20.0 & 64.0 & 84.0 & 60.0 \\
\bottomrule
\end{tabular}
\vspace{-8pt}
\caption{\textbf{Experiments with and without TaskMoE.} TaskMoE not only improves performance on trained tasks but also enhances generalization to unseen tasks. Here, PID, TT, PGC, PMS, and CJ denote \textit{Put Item in Drawer}, \textit{Turn Tap}, \textit{Put Groceries in Cupboard}, \textit{Put Money in Safe}, and \textit{Close Jar}, respectively.}
\vspace{-10pt}
\label{tab:womoe}
\end{table}

\begin{table}[t]
\centering
\renewcommand{\arraystretch}{1}
% \rowcolors{2}{lightgray}{white}  % 从第2行开始交替上色
\scriptsize
\setlength{\tabcolsep}{6pt}
\begin{tabular}{c c c c c c c c}
\toprule

$K$ & $r$ & \textbf{PID} & \textbf{TT} & \textbf{PGC} & \textbf{PMS} & \textbf{CJ} & \textbf{Avg. SR (\%)} $\uparrow$\\ 
\midrule
2 & 0.75$\sim$1.3  & 12.0 & 92.0 & 0.0 & 12.0 & 20.0 & 27.2 \\
3 & 0.75$\sim$1.3  & 32.0 & 92.0 & 16.0 &  32.0& 76.0 & 49.6 \\
4 & 0.75$\sim$1.3  & 52.0 & 100.0 & 4.0 & 68.0 & 52.0 & 55.2 \\
3 & 0.60$\sim$1.56 & 16.0 & 96.0 & 8.0 & 44.0 & 80.0 & 48.8 \\
3 & 0.90$\sim$1.04 & 44.0 & 100.0 & 20.0 & 36.0 & 80.0 & 56.0 \\
\bottomrule
\end{tabular}
\vspace{-8pt}
\caption{Performance under different view numbers $K$ and camera radial constraints $r$, which $\Delta(r_{min}, r_{max})=\pm20\%$. Task PID $\to$ \textit{Put item in drawer}, TT $\to$ \textit{Turn tap}, PGC $\to$ \textit{Put groceries in cupboard}, PMS $\to$ \textit{Put money in safe}, CJ $\to$ \textit{Close jar}.}
% \zhouxia{It is hard to analyze why we used $k=3$ and $0.75 \sim 1.3$ with these resutls.}}
\vspace{-15pt}
\label{tab:camera_constraints}
\end{table}

\noindent\textbf{Setting of $K$ and $r$.}
We conducted ablation studies on the camera radial constraint $r$ and the number of views $K$. Compared to the baseline range ($0.75\sim1.3$)\text{m}, a tighter constraint ($0.90 \sim 1.04$)\text{m} improved performance, raising the average success rate from $49.6\%$ to $56.0\%$ (Table~\ref{tab:camera_constraints}). A looser constraint ($0.60 \sim 1.56$)\text{m} slightly reduced performance ($48.8\%$), indicating that a precise viewing distance range enhances task execution. Increasing the number of views from 2 to 4 under the baseline radial constraint consistently improved performance, with average success rates rising from $27.2\%$ to $49.6\%$ and $55.2\%$, respectively (Table~\ref{tab:camera_constraints}). This demonstrates that multi-view observation mitigates occlusion and improves spatial perception. Considering computational cost, we chose $K=3$ as a balance.

\section{Conclusion}

Our \ModelAbbr\ framework establishes that dynamic view planning and task-aware representation learning significantly advance robotic manipulation capabilities. The MVEP's viewpoint optimization effectively overcomes occlusion limitations in fixed-viewpoint systems, while TaskMoE's specialized feature extraction mitigates multi-task interference. These contributions enhance performance across diverse manipulation challenges and enable meaningful generalization to novel tasks. However, the framework has some limitations: multi-view re-rendering slightly increases inference latency, and reliance on accurate global point clouds makes it challenging to handle reflective or transparent objects. Future work could integrate multi-sensor data and domain adaptation to improve real-world robustness.

% However, the framework has some limitations: there is a slight increase in inference latency due to multi-view re-rendering. It relies on accurate global point clouds, making it difficult to handle reflective/transparent objects in real-world environments. Future work could explore multi-sensor integration and domain adaptation techniques to enhance real-world robustness. The modular design of \ModelAbbr\ components offers promising pathways for extending these capabilities to broader manipulation scenarios.

% 添加对加速的总结，将限制移到这里

% 添加ACKNOWLEDGMENTS
\section*{Acknowledgments}

This work is supported by National Key Research and Development Program of China (2024YFE0203100), in part by the National Natural Science Foundation of China under Grant No.62572498 and No.62536010, in part by the open research fund of Pengcheng Laboratory under Grant 2025KF1B0050, in part by the Guangdong Basic and Applied Basic Research Foundation under Grant NO. 2025A1515011874. We also thank the National Supercomputer Center in Guangzhou for computational support.

% \clearpage
{
    \small
    \bibliographystyle{ieeenat_fullname}
    \bibliography{main}

@String(ICLR = {Int. Conf. Learn. Represent.})

@String(AAAI = {AAAI})

@string(JAIR = {Journal of Artificial Intelligence Research})

@String(ICLR  = {ICLR})

@String(CoRL = {Conference on Robot Learning})

@String(IJRR = {The International Journal of Robotics Research})

@inproceedings{shazeer2017outrageously,
  title={Outrageously Large Neural Networks: The Sparsely-Gated Mixture-of-Experts Layer},
  author={Shazeer, Noam and Mirhoseini, Azalia and Maziarz, Krzysztof and Davis, Andy and Le, Quoc and Hinton, Geoffrey and Dean, Jeff},
  booktitle={ICLR},
  year={2017}
}

@inproceedings{wangsparse,
    title={Sparse Diffusion Policy: A Sparse, Reusable, and Flexible Policy for Robot Learning},
  author={Wang, Yixiao and Zhang, Yifei and Huo, Mingxiao and Tian, Thomas and Zhang, Xiang and Xie, Yichen and Xu, Chenfeng and Ji, Pengliang and Zhan, Wei and Ding, Mingyu and others},
  booktitle={Conference on Robot Learning},
  pages={649--665},
  year={2025},
  organization={PMLR}
}

@article{james2020rlbench,
  title={{RLBench}: The robot learning benchmark \& learning environment},
  author={James, Stephen and Ma, Zicong and Arrojo, David Rovick and Davison, Andrew J},
  journal={IEEE Robotics and Automation Letters},
  year={2020}
}

@article{chi2023diffusion,
  title={{Diffusion Policy}: Visuomotor policy learning via action diffusion},
  author={Chi, Cheng and Xu, Zhenjia and Feng, Siyuan and Cousineau, Eric and Du, Yilun and Burchfiel, Benjamin and Tedrake, Russ and Song, Shuran},
  journal={IJRR},
  year={2023}
}

@article{hou2024diffusion,
  title={Diffusion transformer policy},
  author={Hou, Zhi and Zhang, Tianyi and Xiong, Yuwen and Pu, Hengjun and Zhao, Chengyang and Tong, Ronglei and Qiao, Yu and Dai, Jifeng and Chen, Yuntao},
  journal={arXiv preprint arXiv:2410.15959},
  year={2024}
}

@article{team2024octo,
  title={Octo: An open-source generalist robot policy},
  author={Team, Octo Model and Ghosh, Dibya and Walke, Homer and Pertsch, Karl and Black, Kevin and Mees, Oier and Dasari, Sudeep and Hejna, Joey and Kreiman, Tobias and Xu, Charles and others},
  journal={Proceedings of Robotics: Science and Systems},
  year={2024}
}

@inproceedings{kimopenvla,
  title={{OpenVLA}: An Open-Source Vision-Language-Action Model},
  author={Kim, Moo Jin and Pertsch, Karl and Karamcheti, Siddharth and Xiao, Ted and Balakrishna, Ashwin and Nair, Suraj and Rafailov, Rafael and Foster, Ethan P and Sanketi, Pannag R and Vuong, Quan and others},
  booktitle={CoRL},
  year={2024}
}

@inproceedings{ze20243d,
  title={3D Diffusion Policy: Generalizable Visuomotor Policy Learning via Simple 3D Representations},
  author={Ze, Yanjie and Zhang, Gu and Zhang, Kangning and Hu, Chenyuan and Wang, Muhan and Xu, Huazhe},
  booktitle={ICRA},
  year={2024}
}

@article{wang2024poco,
  title={{PoCo}: Policy Composition from and for Heterogeneous Robot Learning},
  author={Wang, Lirui and Zhao, Jialiang and Du, Yilun and Adelson, Edward H and Tedrake, Russ},
  journal={CoRR},
  year={2024}
}

@inproceedings{shridhar2022cliport,
  title={{CLIPort}: What and where pathways for robotic manipulation},
  author={Shridhar, Mohit and Manuelli, Lucas and Fox, Dieter},
  booktitle={CoRL},
  year={2022}
}

@inproceedings{nair2022learning,
  title={Learning language-conditioned robot behavior from offline data and crowd-sourced annotation},
  author={Nair, Suraj and Mitchell, Eric and Chen, Kevin and Savarese, Silvio and Finn, Chelsea and others},
  booktitle={CoRL},
  year={2022}
}

@article{rt-2,
  title={{RT-2}: Vision-language-action models transfer web knowledge to robotic control},
  author={Brohan, Anthony and Brown, Noah and Carbajal, Justice and Chebotar, Yevgen and Chen, Xi and Choromanski, Krzysztof and Ding, Tianli and Driess, Danny and Dubey, Avinava and Finn, Chelsea and others},
  journal={CoRL},
  year={2023}
}

@article{cheang2024gr,
  title={GR-2: A generative video-language-action model with web-scale knowledge for robot manipulation},
  author={Cheang, Chi-Lam and Chen, Guangzeng and Jing, Ya and Kong, Tao and Li, Hang and Li, Yifeng and Liu, Yuxiao and Wu, Hongtao and Xu, Jiafeng and Yang, Yichu and others},
  journal={arXiv preprint arXiv:2410.06158},
  year={2024}
}

@article{zhang2025autoregressive,
  title={Autoregressive action sequence learning for robotic manipulation},
  author={Zhang, Xinyu and Liu, Yuhan and Chang, Haonan and Schramm, Liam and Boularias, Abdeslam},
  journal={IEEE Robotics and Automation Letters},
  year={2025}
}

@article{li2025bridgevla,
  title={{BridgeVLA}: Input-Output Alignment for Efficient 3D Manipulation Learning with Vision-Language Models},
  author={Li, Peiyan and Chen, Yixiang and Wu, Hongtao and Ma, Xiao and Wu, Xiangnan and Huang, Yan and Wang, Liang and Kong, Tao and Tan, Tieniu},
  journal={arXiv preprint arXiv:2506.07961},
  year={2025}
}

@article{lu2025vla,
  title={{VLA-RL}: Towards masterful and general robotic manipulation with scalable reinforcement learning},
  author={Lu, Guanxing and Guo, Wenkai and Zhang, Chubin and Zhou, Yuheng and Jiang, Haonan and Gao, Zifeng and Tang, Yansong and Wang, Ziwei},
  journal={arXiv preprint arXiv:2505.18719},
  year={2025}
}

@article{schulman2017proximal,
  title={Proximal policy optimization algorithms},
  author={Schulman, John and Wolski, Filip and Dhariwal, Prafulla and Radford, Alec and Klimov, Oleg},
  journal={arXiv preprint arXiv:1707.06347},
  year={2017}
}

@article{livision,
  title={Vision-language foundation models as effective robot imitators},
  author={Li, Xinghang and Liu, Minghuan and Zhang, Hanbo and Yu, Cunjun and Xu, Jie and Wu, Hongtao and Cheang, Chilam and Jing, Ya and Zhang, Weinan and Liu, Huaping and others},
  journal={ICLR},
  year={2024}
}

@article{mao2024robomatrix,
  title={Robomatrix: A skill-centric hierarchical framework for scalable robot task planning and execution in open-world},
  author={Mao, Weixin and Zhong, Weiheng and Jiang, Zhou and Fang, Dong and Zhang, Zhongyue and Lan, Zihan and Li, Haosheng and Jia, Fan and Wang, Tiancai and Fan, Haoqiang and others},
  journal={arXiv preprint arXiv:2412.00171},
  year={2024}
}

@article{haldar2024baku,
  title={Baku: An efficient transformer for multi-task policy learning},
  author={Haldar, Siddhant and Peng, Zhuoran and Pinto, Lerrel},
  journal={Advances in Neural Information Processing Systems},
  volume={37},
  pages={141208--141239},
  year={2024}
}

@inproceedings{dalallocal,
  title={Local Policies Enable Zero-shot Long-horizon Manipulation},
  author={Dalal, Murtaza and Liu, Min and Talbott, Walter and Chen, Chen and Pathak, Deepak and Zhang, Jian and Salakhutdinov, Russ},
  booktitle={2nd CoRL Workshop on Learning Effective Abstractions for Planning},
  year={2024},
}

@inproceedings{deisenroth2014multi,
  title={Multi-task policy search for robotics},
  author={Deisenroth, Marc Peter and Englert, Peter and Peters, Jan and Fox, Dieter},
  booktitle={2014 IEEE international conference on robotics and automation (ICRA)},
  pages={3876--3881},
  year={2014},
  organization={IEEE}
}

@inproceedings{ze2023gnfactor,
  title={Gnfactor: Multi-task real robot learning with generalizable neural feature fields},
  author={Ze, Yanjie and Yan, Ge and Wu, Yueh-Hua and Macaluso, Annabella and Ge, Yuying and Ye, Jianglong and Hansen, Nicklas and Li, Li Erran and Wang, Xiaolong},
  booktitle={Conference on robot learning},
  pages={284--301},
  year={2023},
  organization={PMLR}
}

@inproceedings{ma2024hierarchical,
  title={Hierarchical diffusion policy for kinematics-aware multi-task robotic manipulation},
  author={Ma, Xiao and Patidar, Sumit and Haughton, Iain and James, Stephen},
  booktitle={Proceedings of the IEEE/CVF Conference on Computer Vision and Pattern Recognition},
  pages={18081--18090},
  year={2024}
}

@inproceedings{tang2025deep,
  title={Deep reinforcement learning for robotics: A survey of real-world successes},
  author={Tang, Chen and Abbatematteo, Ben and Hu, Jiaheng and Chandra, Rohan and Mart{\'\i}n-Mart{\'\i}n, Roberto and Stone, Peter},
  booktitle={Proceedings of the AAAI Conference on Artificial Intelligence},
  volume={39},
  number={27},
  pages={28694--28698},
  year={2025}
}

@article{hu2024flare,
  title={Flare: Achieving masterful and adaptive robot policies with large-scale reinforcement learning fine-tuning},
  author={Hu, Jiaheng and Hendrix, Rose and Farhadi, Ali and Kembhavi, Aniruddha and Mart{\'\i}n-Mart{\'\i}n, Roberto and Stone, Peter and Zeng, Kuo-Hao and Ehsani, Kiana},
  journal={arXiv preprint arXiv:2409.16578},
  year={2024}
}

@article{guo2025improving,
  title={Improving vision-language-action model with online reinforcement learning},
  author={Guo, Yanjiang and Zhang, Jianke and Chen, Xiaoyu and Ji, Xiang and Wang, Yen-Jen and Hu, Yucheng and Chen, Jianyu},
  journal={arXiv preprint arXiv:2501.16664},
  year={2025}
}

@article{rajeswaran2017learning,
  title={Learning complex dexterous manipulation with deep reinforcement learning and demonstrations},
  author={Rajeswaran, Aravind and Kumar, Vikash and Gupta, Abhishek and Vezzani, Giulia and Schulman, John and Todorov, Emanuel and Levine, Sergey},
  journal={arXiv preprint arXiv:1709.10087},
  year={2017}
}

@inproceedings{uchendu2023jump,
  title={Jump-start reinforcement learning},
  author={Uchendu, Ikechukwu and Xiao, Ted and Lu, Yao and Zhu, Banghua and Yan, Mengyuan and Simon, Jos{\'e}phine and Bennice, Matthew and Fu, Chuyuan and Ma, Cong and Jiao, Jiantao and others},
  booktitle={International Conference on Machine Learning},
  pages={34556--34583},
  year={2023},
  organization={PMLR}
}

@inproceedings{luo2024serl,
  title={Serl: A software suite for sample-efficient robotic reinforcement learning},
  author={Luo, Jianlan and Hu, Zheyuan and Xu, Charles and Tan, You Liang and Berg, Jacob and Sharma, Archit and Schaal, Stefan and Finn, Chelsea and Gupta, Abhishek and Levine, Sergey},
  booktitle={2024 IEEE International Conference on Robotics and Automation (ICRA)},
  pages={16961--16969},
  year={2024},
  organization={IEEE}
}

@article{khetarpal2022toward,
  title={Towards continual reinforcement learning: A review and perspectives},
  author={Khetarpal, Khimya and Riemer, Matthew and Rish, Irina and Precup, Doina},
  journal=JAIR,
  volume={75},
  pages={1401--1476},
  year={2022}
}

@inproceedings{fang2020graspnet,
  title={Graspnet-1billion: A large-scale benchmark for general object grasping},
  author={Fang, Hao-Shu and Wang, Chenxi and Gou, Minghao and Lu, Cewu},
  booktitle={Proceedings of the IEEE/CVF conference on computer vision and pattern recognition},
  pages={11444--11453},
  year={2020}
}

@article{bu2024towards,
  title={Towards synergistic, generalized, and efficient dual-system for robotic manipulation},
  author={Bu, Qingwen and Li, Hongyang and Chen, Li and Cai, Jisong and Zeng, Jia and Cui, Heming and Yao, Maoqing and Qiao, Yu},
  journal={arXiv preprint arXiv:2410.08001},
  year={2024}
}

@inproceedings{deng2020self,
  title={Self-supervised 6d object pose estimation for robot manipulation},
  author={Deng, Xinke and Xiang, Yu and Mousavian, Arsalan and Eppner, Clemens and Bretl, Timothy and Fox, Dieter},
  booktitle={2020 IEEE International Conference on Robotics and Automation (ICRA)},
  pages={3665--3671},
  year={2020},
  organization={IEEE}
}

@article{prianto2020path,
  title={Path planning for multi-arm manipulators using deep reinforcement learning: Soft actor--critic with hindsight experience replay},
  author={Prianto, Evan and Kim, MyeongSeop and Park, Jae-Han and Bae, Ji-Hun and Kim, Jung-Su},
  journal={Sensors},
  volume={20},
  number={20},
  pages={5911},
  year={2020},
  publisher={MDPI}
}

@article{zhao2025deep,
  title={Deep reinforcement learning trajectory planning for robotic manipulator based on simulation-efficient training},
  author={Zhao, Bin and Wu, Yao and Wu, Chengdong and Sun, Ruohuai},
  journal={Scientific Reports},
  volume={15},
  number={1},
  pages={8286},
  year={2025},
  publisher={Nature Publishing Group UK London}
}

@article{bu2025univla,
  title={Univla: Learning to act anywhere with task-centric latent actions},
  author={Bu, Qingwen and Yang, Yanting and Cai, Jisong and Gao, Shenyuan and Ren, Guanghui and Yao, Maoqing and Luo, Ping and Li, Hongyang},
  journal={arXiv preprint arXiv:2505.06111},
  year={2025}
}

@article{cen2025worldvla,
  title={WorldVLA: Towards Autoregressive Action World Model},
  author={Cen, Jun and Yu, Chaohui and Yuan, Hangjie and Jiang, Yuming and Huang, Siteng and Guo, Jiayan and Li, Xin and Song, Yibing and Luo, Hao and Wang, Fan and others},
  journal={arXiv preprint arXiv:2506.21539},
  year={2025}
}

@article{liu2025hybridvla,
  title={Hybridvla: Collaborative diffusion and autoregression in a unified vision-language-action model},
  author={Liu, Jiaming and Chen, Hao and An, Pengju and Liu, Zhuoyang and Zhang, Renrui and Gu, Chenyang and Li, Xiaoqi and Guo, Ziyu and Chen, Sixiang and Liu, Mengzhen and others},
  journal={arXiv preprint arXiv:2503.10631},
  year={2025}
}

@article{lin2025onetwovla,
  title={OneTwoVLA: A Unified Vision-Language-Action Model with Adaptive Reasoning},
  author={Lin, Fanqi and Nai, Ruiqian and Hu, Yingdong and You, Jiacheng and Zhao, Junming and Gao, Yang},
  journal={arXiv preprint arXiv:2505.11917},
  year={2025}
}

@article{rt-1,
  title={{RT-1}: Robotics Transformer for Real-World Control at Scale},
  author={Brohan, Anthony and Brown, Noah and Carbajal, Justice and Chebotar, Yevgen and Dabis, Joseph and Finn, Chelsea and Gopalakrishnan, Keerthana and Hausman, Karol and Herzog, Alexander and Hsu, Jasmine and others},
  journal={Robotics: Science and Systems XIX},
  year={2023},
  publisher={Robotics: Science and Systems Foundation}
}

@article{black2410pi0,
  title={$\pi$0: A vision-language-action flow model for general robot control},
  author={Black, Kevin and Brown, Noah and Driess, Danny and Esmail, Adnan and Equi, Michael and Finn, Chelsea and Fusai, Niccolo and Groom, Lachy and Hausman, Karol and Ichter, Brian and others},
  journal={arXiv preprint ARXIV.2410.24164},
year={2024},
}

@article{welford1962note,
  title={Note on a method for calculating corrected sums of squares and products},
  author={Welford, Barry Payne},
  journal={Technometrics},
  volume={4},
  number={3},
  pages={419--420},
  year={1962},
  publisher={Taylor \& Francis}
}

@article{ACT,
  title={Learning fine-grained bimanual manipulation with low-cost hardware},
  author={Zhao, Tony Z and Kumar, Vikash and Levine, Sergey and Finn, Chelsea},
  journal={RSS},
  year={2023}
}

@inproceedings{goyal2024rvt2,
  title={RVT-2: Learning Precise Manipulation from Few Demonstrations},
  author={Goyal, Ankit and Blukis, Valts and Xu, Jie and Guo, Yijie and Chao, Yu-Wei and Fox, Dieter},
  booktitle={RSS 2024 Workshop: Data Generation for Robotics},
  year={2024}
}

@article{goyal2023rvt,
  title={RVT: Robotic View Transformer for 3D Object Manipulation},
  author={Goyal, Ankit and Xu, Jie and Guo, Yijie and Blukis, Valts and Chao, Yu-Wei and Fox, Dieter},
  journal={arXiv preprint arXiv:2306.14896},
  year={2023}
}

@inproceedings{shridhar2023perceiver,
  title={Perceiver-actor: A multi-task transformer for robotic manipulation},
  author={Shridhar, Mohit and Manuelli, Lucas and Fox, Dieter},
  booktitle={Conference on Robot Learning},
  pages={785--799},
  year={2023},
  organization={PMLR}
}

@article{ke20243ddiffactor,
  title={3d diffuser actor: Policy diffusion with 3d scene representations},
  author={Ke, Tsung-Wei and Gkanatsios, Nikolaos and Fragkiadaki, Katerina},
  journal={arXiv preprint arXiv:2402.10885},
  year={2024}
}

@inproceedings{james2022coarse,
  title={Coarse-to-fine q-attention: Efficient learning for visual robotic manipulation via discretisation},
  author={James, Stephen and Wada, Kentaro and Laidlow, Tristan and Davison, Andrew J},
  booktitle={Proceedings of the IEEE/CVF Conference on Computer Vision and Pattern Recognition},
  pages={13739--13748},
  year={2022}
}

@inproceedings{Film,
  title={Film: Visual reasoning with a general conditioning layer},
  author={Perez, Ethan and Strub, Florian and De Vries, Harm and Dumoulin, Vincent and Courville, Aaron},
  booktitle={AAAI},
  year={2018}
}

@article{lookatmodel,
  title={Transporter networks: Rearranging the visual world for robotic manipulation},
  author={Zeng, A and Florence, P and Tompson, J and Welker, S and Chien, J and Attarian, M and Armstrong, T and Krasin, I},
  journal={RSS},
  year={2021}
}

@inproceedings{guhur2023instruction,
  title={Instruction-driven history-aware policies for robotic manipulations},
  author={Guhur, Pierre-Louis and Chen, Shizhe and Pinel, Ricardo Garcia and Tapaswi, Makarand and Laptev, Ivan and Schmid, Cordelia},
  booktitle={Conference on Robot Learning},
  pages={175--187},
  year={2023},
  organization={PMLR}
}

@inproceedings{gervet2023act3d,
  title={Act3D: 3D Feature Field Transformers for Multi-Task Robotic Manipulation},
  author={Gervet, Theophile and Xian, Zhou and Gkanatsios, Nikolaos and Fragkiadaki, Katerina},
  booktitle={Conference on Robot Learning},
  pages={3949--3965},
  year={2023},
  organization={PMLR}
}

@inproceedings{ke20253d,
  title={3D Diffuser Actor: Policy Diffusion with 3D Scene Representations},
  author={Ke, Tsung-Wei and Gkanatsios, Nikolaos and Fragkiadaki, Katerina},
  booktitle={Conference on Robot Learning},
  pages={1949--1974},
  year={2025},
  organization={PMLR}
}

@inproceedings{chen2023polarnet,
  title={PolarNet: 3D Point Clouds for Language-Guided Robotic Manipulation},
  author={Chen, Shizhe and Garcia, Ricardo and Schmid, Cordelia and Laptev, Ivan},
  booktitle={7th Conference on Robot Learning (CoRL 2023)},
  year={2023}
}

@article{liu2025aligning,
  title={Aligning cyber space with physical world: A comprehensive survey on embodied ai},
  author={Liu, Yang and Chen, Weixing and Bai, Yongjie and Liang, Xiaodan and Li, Guanbin and Gao, Wen and Lin, Liang},
  journal={IEEE/ASME Transactions on Mechatronics},
  year={2025},
  publisher={IEEE}
}

@inproceedings{pumacay2024colosseum,
  title={The COLOSSEUM: A Benchmark for Evaluating Generalization for Robotic Manipulation},
  author={Pumacay, Wilbert and Singh, Ishika and Duan, Jiafei and Krishna, Ranjay and Thomason, Jesse and Fox, Dieter},
  booktitle={RSS 2024 Workshop: Data Generation for Robotics},
  year={2024}
}
}


\clearpage
% \appendix
{ % 开始局部作用域
\centering
\section*{Appendix}
} % 结束局部作用域，对齐方式恢复默认
\setcounter{section}{1}
In this supplementary material, we provide additional details and results of our proposed \ModelAbbr\ to complement the main paper. The content is organized as follows:
\begin{itemize}
    \item \textbf{Appendix A:} Real-world experimental details. Corresponding \textbf{demo videos} of the results are also included in the supplementary materials.
    % web project page \href{https://hcplab-sysu.github.io/\ModelAbbr\}{\ModelAbbr\} for a more intuitive understanding.
    \item \textbf{Appendix B:} Additional design details of \ModelAbbr, including TaskMoE, point cloud aggregation, camera pose parameterization, and the proposed pseudo-environment interaction mechanism.
    \item \textbf{Appendix C:} More simulation experimental details and results.
    \item \textbf{Appendix D:} Implementation details and information about the RLBench~\cite{james2020rlbench} and RLBench-OG. Additionally, we provide more visualized results, with corresponding \textbf{demo videos} included in the supplementary materials.
    \item \textbf{Appendix E:} Multi-view re-rendering visualization results.
    % \item \textbf{Appendix D:} Discuss and analyze the limitations of our \ModelAbbr\ model.
\end{itemize}

\section*{Appendix A. Real-World Experimental Details}
\label{appendix:real_world_experiments}

\begin{figure}[th]
    \centering
\includegraphics[width=1\linewidth]{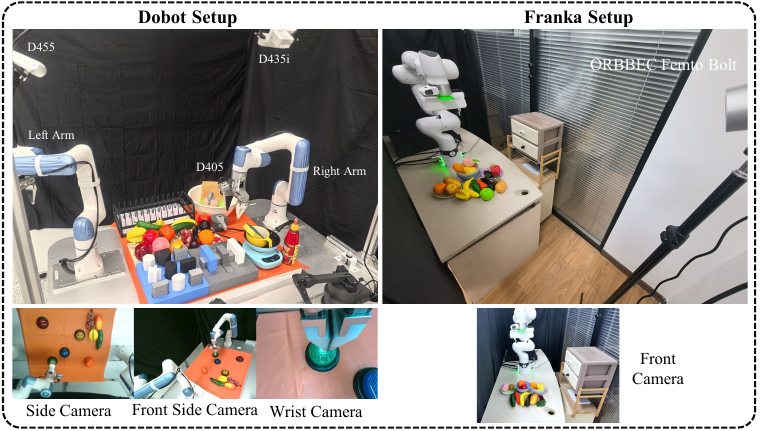}
\vspace{-20pt}
    \caption{Real-World Environment Setup.}
    \vspace{-10pt}
    \label{fig:hw-env}
\end{figure}

To validate the generalizability of the proposed \ModelAbbr\ in real-world deployment scenarios, we conducted experiments on two distinct robotic platforms: the Dobot Nova 2 and the Franka Research 3, employing differentiated camera configurations (multi-view and single-view) to comprehensively validate the scenarios. For the Dobot Nova 2, a multi-view setup consisting of three perspectives (side, front, and wrist-mounted) was used, whereas for the Franka robot, only a front view was utilized. Experimental results confirm that our \ModelAbbr\ can be successfully deployed on both platforms and outperforms both Diffusion Policy and ARP. Figure \ref{fig:hw-env} illustrates the real-world deployment environments and the corresponding images from each viewpoint.

Across five real-world tasks—\textit{Pick Grape}, \textit{Stack Bowls}, \textit{Push Buttons}, \textit{Collect Fruits}, and \textit{Put Item In Drawer}—our \ModelAbbr\ achieves strong performance. Notably, in the \textit{pick\_grape} task, it successfully executed the manipulation even when the target was partially occluded in one viewpoint, demonstrating the advantage of our model in global perception capabilities. Corresponding demonstration videos are provided to facilitate a more intuitive understanding of the real-world outcomes.

Furthermore, to further evaluate the executive capability of our method across various types of tasks, we extended the experiments on the \textit{Franka Research 3} platform to include ten additional tasks. These tasks encompass articulated object manipulation, deformable object manipulation, among others. Experimental results are summarized in Table \ref{tab:other_task_comparison_in_real_world}, which indicate that our \ModelAbbr\ achieves high success rates, highlighting its adaptability and effectiveness across a diverse set of tasks.

\begin{table*}[h]
\centering
\scriptsize
\resizebox{1.0\linewidth}{!}{%
\small
\setlength{\tabcolsep}{2.0pt}
\begin{tabular}{llc|cccccccccc}
\toprule
\textbf{Arm} & \textbf{Method} & \textbf{Avg. SR (\%)} $\uparrow$ &
\textbf{Pick Place} &
\textbf{Stack Bowls} &
\textbf{Push Buttons} &
\textbf{Collect Fruits} &
\textbf{Put Item In Drawer} &
\textbf{Rotate Handle} &
\textbf{Fold Towel} &
\textbf{Open Lip} &
\textbf{Unscrew Bottle} &
\textbf{Reach Drag} \\
\midrule
\multirow{2}{*}{Franka}
% Franka
% & Diffusion Policy 
% & -- 
% & 90.0 & 50.0 & 60.0 & 40.0 & 50.0 
% & -- & -- & -- & -- & -- \\

& ARP              
& 62.0\%
& 10/10 & 7/10 & 8/10 & 5/10 & 6/10
& 3/10 & 5/10 & 7/10 & 3/10 & 8/10 \\

& \ModelAbbr\ (Ours)
& 70.0\%
& 10/10 & 7/10 & 9/10 & 6/10 & 7/10
& 3/10 & 5/10 & 9/10 & 5/10 & 9/10 \\
\bottomrule
\end{tabular}
}
\vspace{-8pt}
% \caption{Comparison of task success rates for Franka robots across more manipulation tasks.}
\caption{Task success rates for Franka robots across more manipulation tasks.}
\vspace{-15pt}
\label{tab:other_task_comparison_in_real_world}
\end{table*}

\subsection*{Real-World Robustness and Generalization Testing.}
% \begin{table}[H]
% \centering
% \scriptsize
% \vspace{-5pt}
% \begin{tabular}{llcccccc}
% \toprule
% \textbf{Arm} & \textbf{Method} & \textbf{Seen} & \textbf{Inst.} & \textbf{Bkg.} & \textbf{Obj.} & \textbf{Occl.} & \textbf{Illum.} \\
% \midrule
% \multirow{2}{*}{Dobot} & Diffusion Policy & 90.0 & 80.0 & 70.0 & 60.0 & 10.0 & 10.0 \\
% & \ModelAbbr\ (Ours) & 100.0 & 100.0 & 90.0 & 90.0 & 20.0 & 30.0 \\
% \bottomrule
% \end{tabular}
% \vspace{-10pt}
% \caption{Performance comparison under different configurations on the Dobot robot. Abbreviations: Inst. (Unseen Instance), Bkg. (Unseen Background), Obj. (Unseen Object), Occl. (Heavy Occlusion), Illum. (Illumination Variation).}
% \vspace{-10pt}
% \label{tab:cfg_comparison}
% \end{table}

\begin{table}[H]
\centering
\vspace{-10pt}
\scriptsize
\resizebox{1.005\linewidth}{!}{%
\small
\setlength{\tabcolsep}{2.0pt}
\begin{tabular}{llc|cccccc}
\toprule
\textbf{Arm} & \textbf{Method} & \textbf{Avg. SR (\%)} $\uparrow$ & \textbf{Seen} & \textbf{Inst.} & \textbf{Bkg.} & \textbf{Obj.} & \textbf{Occl.} & \textbf{Illum.} \\
\midrule
\multirow{2}{*}{Dobot} 
& Diffusion Policy & 53.3 & 90.0 & 80.0 & 70.0 & 60.0 & 10.0 & 10.0 \\
& \ModelAbbr\ (Ours) & \textbf{71.7} & 100.0 & 100.0 & 90.0 & 90.0 & 20.0 & 30.0 \\
\bottomrule
\end{tabular}
}
\vspace{-10pt}
\caption{Performance comparison under different configurations on the Dobot robot. Abbreviations: Inst. (Unseen Instance), Bkg. (Unseen Background), Obj. (Unseen Object), Occl. (Heavy Occlusion), Illum. (Illumination Variation).}
\vspace{-10pt}
\label{tab:cfg_comparison}
\end{table}

We conduct robustness testing on the real-world \textit{Pick Grape} task on the \textit{Dobot nova 2} robot, evaluating under various conditions including unseen instances, unseen backgrounds, unseen objects, heavy occlusion, and illumination variation. Throughout the experiments, heavy occlusion is identified as the primary cause of system failures. The results demonstrate that \ModelAbbr\ exhibits significantly better overall adaptability compared to the Diffusion Policy (DP) baseline. While both methods perform similarly in seen scenarios, \ModelAbbr\ substantially outperforms DP when confronted with unseen backgrounds, unseen objects, and especially under heavy occlusion. The success rates based on 10 trials for each configuration are reported, with quantitative comparisons detailed in Table \ref{tab:cfg_comparison}.

\begin{figure*}[th]
    \centering
    
\includegraphics[width=1\linewidth]{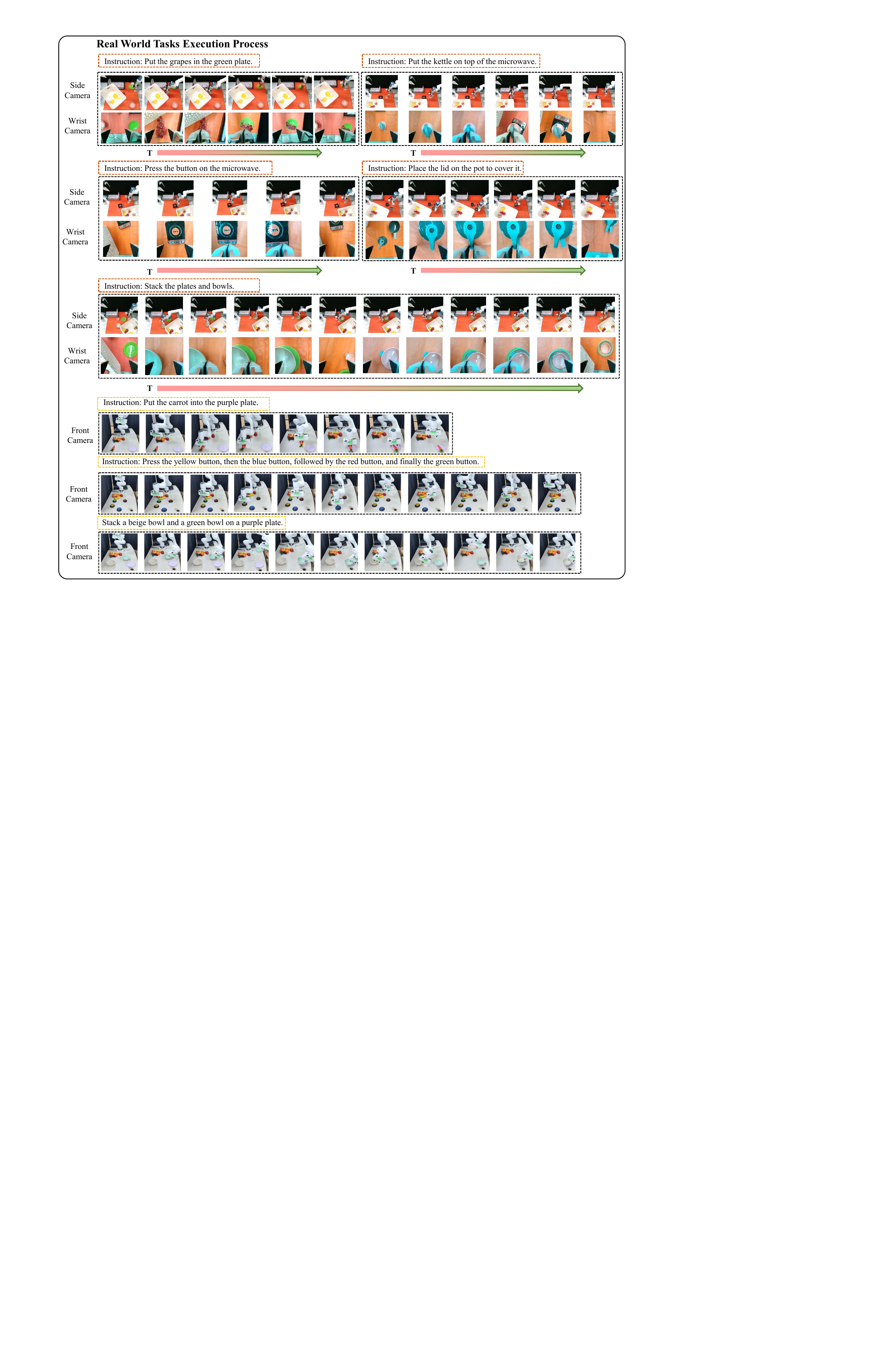}
\vspace{-20pt}
    \caption{Visualization of Dobot and Franka Robots' Execution Processes in Real-World Environments. In the figure, the T-axis represents the time axis, indicating the duration from the start to the end of the task. Each demonstration shows the following elements: side and wrist camera views for the Dobot robot, and a single front view for the Franka robot.}
\vspace{-15pt}
    \label{fig:real_world_res}
\end{figure*}

\subsection*{Details of Real-World Tasks.}

We design real-world tasks to encompass as diverse operation types as possible, including grasping and placing, pressing/pushing, articulated object manipulation, rotation, deformable object manipulation, fine manipulation, tool usage, etc. The design details of each task are elaborated below.

\begin{figure}[H]
    \centering
    \vspace{-10pt}
\includegraphics[width=1\linewidth]{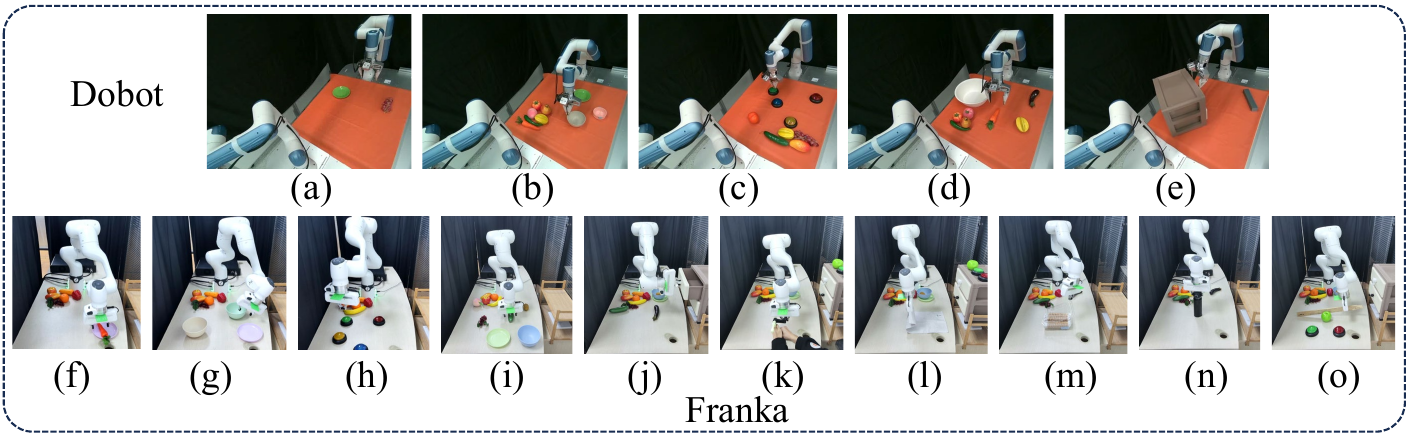}
\vspace{-20pt}
    \caption{Real-World Tasks Setup.}
    \vspace{-16pt}
    \label{fig:real-world-tasks}
\end{figure}

\textbf{Pick and Place.} On the table, there are fruits and a plate. The robot needs to pick up and place the specified fruits. Language instruction: Put the $\mathrm{[FRUIT]}$ into the $\mathrm{[COLOR]}$ plate. As shown in Fig.~\ref{fig:real-world-tasks} (a)(f).

\textbf{Stack Bowls.} There are some bowls and a plate on the table. The robot needs to stack the bowls and plate in a specific order. Language instruction: Stack a $\mathrm{[COLOR]}$ bowl and a $\mathrm{[COLOR]}$ bowl on a $\mathrm{[COLOR]}$ plate. As shown in Fig.~\ref{fig:real-world-tasks} (b)(g).

\textbf{Push Buttons.} There are multiple buttons of different colors on the table. The robot needs to press the buttons in a specific sequence. Language instruction: Press the $\mathrm{[COLOR]}$ button, then the $\mathrm{[COLOR]}$ button, followed by the $\mathrm{[COLOR]}$ button, and finally the $\mathrm{[COLOR]}$ button. As shown in Fig.~\ref{fig:real-world-tasks} (c)(h).

\textbf{Collect Fruits.} There are various fruits on the table. The robot needs to collect the specified fruits into the specific target. Language instruction: Put a $\mathrm{[FRUIT]}$, a $\mathrm{[FRUIT]}$, and an $\mathrm{[FRUIT]}$ into the $\mathrm{[COLOR]}$ $\mathrm{[TARGET]}$. As shown in Fig.~\ref{fig:real-world-tasks} (d)(i).

\textbf{Put Item In Drawer.} There is a cabinet on the table. The robot needs to open the specified drawer, place the target object inside, and then close the drawer. Language instruction: Open the $\mathrm{[Top/Middle/Bottom]}$ drawer and put the $\mathrm{[OBJECT]}$ into it. As shown in Fig.~\ref{fig:real-world-tasks} (e)(j).

\textbf{Rotate Handle.} 
The experimenter holds the handle, and the robot needs to rotate the handle by a certain angle. Language instruction: Rotate the handle clockwise. As shown in Fig.~\ref{fig:real-world-tasks} (k).

\textbf{Fold Towel.} 
There is a towel on the table. The robot needs to pick up one corner of the towel and then fold it. Language instruction: Fold the towel. As shown in Fig.~\ref{fig:real-world-tasks} (l).

\textbf{Open Lip.} 
There is a box on the table. The robot needs to pick up the lid and open it. Language instruction: Open the plastic lid of the biscuit case. As shown in Fig.~\ref{fig:real-world-tasks} (m).

\textbf{Unscrew Bottle.} 
There is a bottle on the table. The robot needs to twist the cap to open the bottle. Language instruction: Unscrew the black bottle. As shown in Fig.~\ref{fig:real-world-tasks} (n).

\textbf{Reach Drag.} 
There is a tool and an object to be manipulated on the table. The robot needs to use the tool to move the target object to the target position. Language instruction: Sweep the green apple into the gap between buttons using wood board. As shown in Fig.~\ref{fig:real-world-tasks} (o).

\section*{Appendix B. More design details of \ModelAbbr}

\subsection*{TaskMoE: Mathematical Principles and Operational Mechanism}

To handle heterogeneous manipulation tasks, we introduce a task-aware Mixture-of-Experts (TaskMoE) that routes visual tokens through (i) a task–instruction–aware gate layer and (ii) a per-gate expert selector. Unlike prior MoE work~\cite{shazeer2017outrageously, wangsparse}, TaskMoE decouples tasks from gates, decouples gates from experts, and adds semantic/entropy regularizers to avoid collapse.

\noindent\textbf{Cross-modal conditioning and FiLM.}
Let $\mathbf{V}\!\in\!\mathbb{R}^{S\times D}$ denote the sequence of visual tokens, where $S$ is the number of spatial tokens and $D$ is the feature dimension. Let $\mathbf{I}\!\in\!\mathbb{R}^{D}$ be the language instruction embedding and $\mathbf{t}\!\in\!\mathbb{R}^{D}$ the learnable embedding of the current task. We first fuse language into vision via cross-modal attention:
\begin{align}
\mathbf{Q} &= \mathbf{V} \mathbf{W}^Q,\quad
\mathbf{K} = \mathbf{I} \mathbf{W}^K,\quad
\mathbf{V}_a = \mathbf{I} \mathbf{W}^V,\\
\mathbf{V}' &= \mathrm{softmax}\!\left(\frac{\mathbf{Q}\mathbf{K}^\top}{\sqrt{d_k}}\right)\mathbf{V}_a,
\end{align}
where $\mathbf{W}^Q,\mathbf{W}^K,\mathbf{W}^V\!\in\!\mathbb{R}^{D\times D}$ are projection matrices and $d_k$ is the key dimension. The attention output $\mathbf{V}'\!\in\!\mathbb{R}^{S\times D}$ is an instruction-attended visual feature. We then modulate channels with FiLM using globally pooled vision and the task embedding (here $\mathrm{GAP}(\cdot)$ is global average pooling over the $S$ tokens):
\begin{align}
\mathbf{c} &= [\mathrm{GAP}(\mathbf{V}');\,\mathbf{t}] \in \mathbb{R}^{2D}, \\
&\quad
\gamma,\beta = \mathrm{MLP}(\mathbf{c}) \in \mathbb{R}^{D}\times\mathbb{R}^{D}, \\
&\quad
\mathbf{V}_{\text{mod}} = \gamma \odot \mathbf{V} + \beta,
\end{align}
where $\gamma$ and $\beta$ are per-channel scaling and bias vectors and $\odot$ denotes element-wise multiplication. The modulated feature $\mathbf{V}_{\text{mod}}\!\in\!\mathbb{R}^{S\times D}$ serves as input to the gating network.

\noindent\textbf{Stage 1: Task-aware gates ($N_G\!\ll\!N_J$).}
We consider $N_J$ tasks and $N_G$ latent gates. Gates represent reusable skill clusters shared across tasks. A shared gate head scores $N_G$ gates for each token and is modulated by a learnable task–gate assignment matrix $\mathbf{A}\in\mathbb{R}^{N_J\times N_G}$:
\begin{align}
\mathbf{p}_g &= \mathrm{softmax}(\mathbf{V}_{\text{mod}} \mathbf{W}_g)\odot \mathrm{softmax}(\mathbf{A}[j,:]),
\end{align}
where $\mathbf{W}_g\!\in\!\mathbb{R}^{D\times N_G}$ is the gate projection and $j\!\in\!\{1,\dots,N_J\}$ is the index of the current task. The matrix $\mathbf{p}_g\!\in\!\mathbb{R}^{S\times N_G}$ contains, for each token, a probability distribution over gates. We select the top-1 gate per token (keeping top-2 for analysis) and apply a per-gate capacity constraint to prevent any gate from absorbing too many tokens.

\noindent\textbf{Stage 2: Per-gate expert selection ($N_E$ arbitrary).}
Given the chosen gate $g\!\in\!\{1,\dots,N_G\}$, we select experts from an independent pool of $N_E$ experts. Each gate has its own expert head:
\begin{align}
\mathbf{z}_e &= \mathbf{V}_{\text{mod}} \mathbf{W}_{\text{exp}}[g] + \mathbf{b}_{\text{exp}}[g],\quad
\mathbf{p}_e = \mathrm{softmax}(\mathbf{z}_e),
\end{align}
where $\mathbf{W}_{\text{exp}}\!\in\!\mathbb{R}^{D\times N_G\times N_E}$ and $\mathbf{b}_{\text{exp}}\!\in\!\mathbb{R}^{N_G\times N_E}$ are the gate–expert projection and bias, and $\mathbf{p}_e\!\in\!\mathbb{R}^{S\times N_E}$ gives, for each token, a probability distribution over experts under gate $g$. We keep top-$k$ experts per token (default $k\!=\!2$), optionally filter the second expert with a threshold/random policy, and enforce an expert capacity across the sequence. The resulting dispatch tensor aggregates these expert weights to route tokens to experts, and expert outputs are combined using the selected mixture weights.

\noindent\textbf{Regularization and loss.}
Let $\mathcal{L}_{\text{task}}$ denote the main imitation learning loss. We add balance, entropy, and semantic regularizers:
\begin{align}
\mathcal{L} &= \mathcal{L}_{\text{task}}
+ \lambda_g \mathcal{L}_{\text{bal-gate}}
+ \lambda_e \mathcal{L}_{\text{bal-exp}} \\
&\quad
- \eta_g H(\mathbf{p}_g)
- \eta_e H(\mathbf{p}_e)
+ \lambda_{\text{sem}} \,\mathcal{L}_{\text{sem}}(\mathbf{A}, \mathbf{I}),
\end{align}
where $\mathcal{L}_{\text{bal-gate}}$ and $\mathcal{L}_{\text{bal-exp}}$ penalize unbalanced gate/expert utilization, $H(\mathbf{p}_g)$ and $H(\mathbf{p}_e)$ are the entropies of the gate and expert distributions (encouraging diverse routing), and $\mathcal{L}_{\text{sem}}(\mathbf{A},\mathbf{I})$ aligns rows of the task–gate matrix $\mathbf{A}$ with the similarity structure of instruction embeddings $\mathbf{I}$ so that semantically related tasks share gates/experts. The scalars $\lambda_g,\lambda_e,\eta_g,\eta_e,\lambda_{\text{sem}}\!\ge\!0$ control the strength of each regularizer. Capacity clipping at both stages further prevents any single gate or expert from monopolizing the traffic.

\noindent\textbf{Scalability and specialization.}
Because gates are decoupled from tasks and from experts, we can set the number of gates $N_G$ and experts $N_E$ independently (e.g., $N_G{=}8$, $N_E{=}16$). The gate layer captures coarse task affinity (each task uses only a few gates), while the expert layer refines routing inside each gate into task-specific specialists. Empirically, increasing $N_E$ leads to finer skill factorization without leaving experts idle, validating the two-stage gate$\rightarrow$expert design with entropy/semantic regularization.

\subsection*{Point Cloud Aggregation.}
\begin{figure}[!t]
    \centering
    \includegraphics[width=0.7\linewidth]{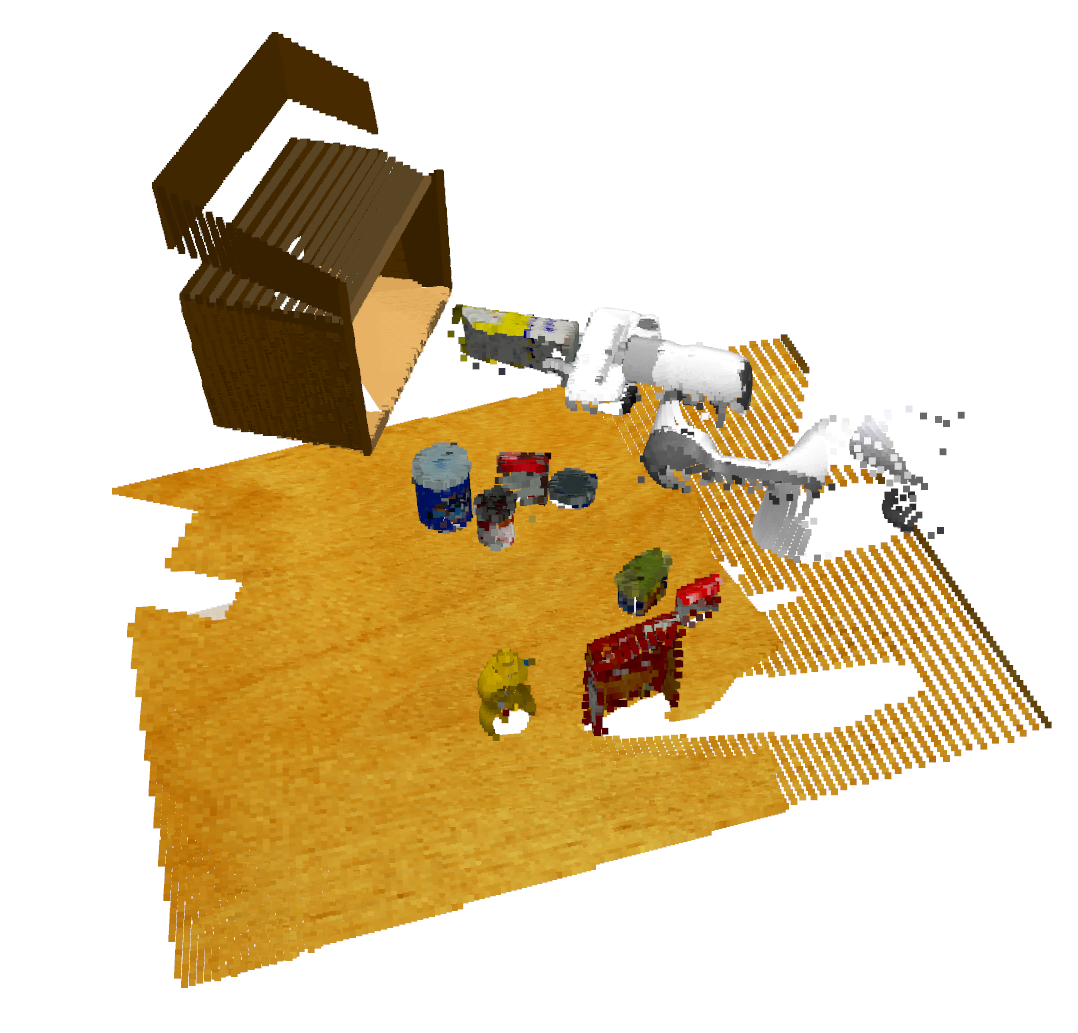}
    \vspace{-15pt}
    \caption{The aggregated full-scene point cloud of the \textit{put\_groceries\_in\_cupboard} task from multiple viewpoints}
    \label{fig:app-pcd}
    \vspace{-10pt}
\end{figure}

In the RLBench dataset experiments, we select RGB-D images from four fixed viewpoints: \textit{front}, \textit{right\_shoulder}, \textit{left\_shoulder}, and \textit{wrist} as inputs. Following the approach of RVT-2~\cite{goyal2024rvt2}, we convert the depth maps from all viewpoints into point clouds in a global coordinate system using the intrinsic and extrinsic camera parameters. Subsequently, we perform multi-view point cloud aggregation and remove redundant points outside the workspace to obtain the final scene point cloud, as illustrated in Fig.~\ref{fig:app-pcd}.

\subsection*{Pseudo-environment Interaction Mechanism.}

To enable efficient policy optimization without physical environment interaction, we design a novel \textit{pseudo-environment} interaction mechanism. As shown in Algorithm~\ref{alg:pseudo-env}. The shadow network $\pi_{\text{shadow}}$ (a frozen copy of the policy before training) processes observation $\mathbf{o}$ to compute reference losses $\mathcal{L}_{\text{ref}}$. MVEP $\pi_{\theta}$ processes $\mathbf{o}$ to generate camera poses $\mathbf{p}$ ($\mathbf{p}$ as policy action), the rendered observation images from multi-view camera poses are fed into TaskMoE and TaskMoE-ARP (Autoregressive Action Policy) to compute the current policy losses $\mathcal{L}_{\text{\ModelAbbr}}$. The reward $r$ is then calculated based on $\mathcal{L}_{\text{ref}}$, $\mathcal{L}_{\text{\ModelAbbr}}$, and $\mathbf{p}$. Finally, the experience tuple $(\mathbf{o}, r, \mathbf{p}, \log \pi_{\theta_{\text{old}}}(\mathbf{p}|\mathbf{o}), V_{\theta_{\text{old}}}(\mathbf{o}))$ is stored in the replay buffer $\mathcal{B}$ for policy updates. This approach enables batch environment interaction, leveraging existing demonstration data to improve policies and enhance data utilization efficiency.

\begin{algorithm}[!t]
\caption{Pseudo-Environment Interaction}
\label{alg:pseudo-env}
\begin{algorithmic}[1]
\State \textbf{Input:} Current observation $\mathbf{o}$, shadow network $\pi_{\text{shadow}}$
\State \textbf{Output:} Transition tuple $\tau$
\State \textit{\# Compute reference loss}
\State $\mathcal{L}_{\text{ref}} \gets \pi_{\text{shadow}}(\mathbf{o})$
\State \textit{\# Generate camera poses \& compute current model loss using \ModelAbbr}
\State $\mathcal{L}_{\text{\ModelAbbr}}, \mathbf{p}, \log \pi_{\theta_{\text{old}}}(\mathbf{p}|\mathbf{o}), V_{\theta_{\text{old}}}(\mathbf{o}) \gets \pi_{\theta_{\text{old}}}(\mathbf{o})$
\State \textit{\# Compute reward}
\State $r \gets \text{RewardCalculator}(\mathcal{L}_{\text{ref}}, \mathcal{L}_{\text{\ModelAbbr}}, \mathbf{p})$
\State \textit{\# Construct transition tuple}
\State \textbf{return} $(\mathbf{o}, \mathbf{p}, r, \log \pi_{\theta_{\text{old}}}(\mathbf{p}|\mathbf{o}), V_{\theta_{\text{old}}}(\mathbf{o}))$
\end{algorithmic}
\end{algorithm}

\subsection*{Camera Pose Parameterization.}

\label{sec:camera-param}
The camera pose is mathematically represented using a \textbf{look-at model} defined by a 5-dimensional parameter vector $\mathbf{p}^{i}  = (\theta^{i}, \phi^{i}, r^{i}, \theta_\text{up}^{i}, \phi_\text{up}^{i}) \in \mathbb{R}^5$($i \in [0, K-1]$, $K$ is number of camera viewpoint). This representation decouples camera position from orientation through spherical coordinates:

\begin{align}
\mathbf{t^{i}} &= 
\begin{bmatrix}
x^{i} \\ y^{i} \\ z^{i}
\end{bmatrix} = 
\begin{bmatrix}
r^{i} \sin\theta^{i} \cos\phi^{i} \\
r^{i} \sin\theta^{i} \sin\phi^{i} \\
r^{i} \cos\theta^{i}
\end{bmatrix}
\label{eq:position}
\end{align}
where $\mathbf{t}^{i}$ denotes the camera center position. The viewing direction is intrinsically defined as $\mathbf{v}_\text{look}^{i} = -\mathbf{t}^{i}/\|\mathbf{t}^{i}\|_2$ toward the \textbf{origin}. The up-vector orientation is parameterized as:

\begin{align}
\mathbf{v}_\text{up}^{i} = 
\begin{bmatrix}
\sin\theta_\text{up}^{i} \cos\phi_\text{up}^{i} \\
\sin\theta_\text{up}^{i} \sin\phi_\text{up}^{i} \\
\cos\theta_\text{up}^{i}
\end{bmatrix}
\label{eq:up_vector}
\end{align}

The final camera matrix is constructed via Gram-Schmidt orthogonalization between $\mathbf{v}_\text{look}^{i}$ and $\mathbf{v}_\text{up}^{i}$.

\begin{figure*}[h!]
    % \centering
    \vspace{-5pt}
\includegraphics[width=1\linewidth]{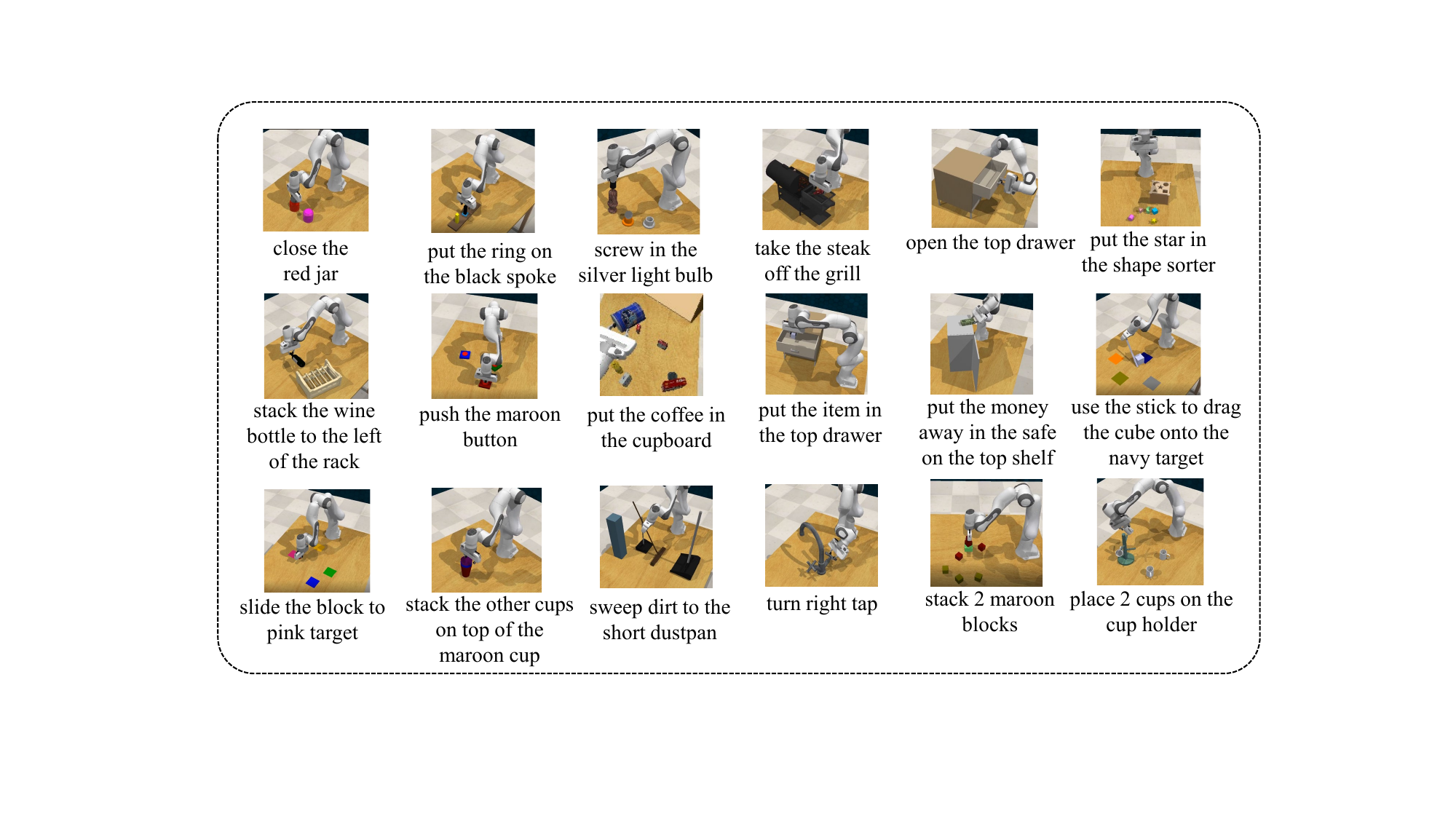}
\vspace{-25pt}
    \caption{Demonstrations of 18 RLBench tasks and their corresponding language instructions.}
    \vspace{-15pt}
    \label{fig:app-rlbench}
\end{figure*}

\section*{Appendix C. More Experimental Details and Resutls}

\subsection*{Baseline Details.}

We compare with eleven state-of-the-art baselines on RLBench, RLbench-OG and Real-world, including both 2D and 3D approaches. For 2D methods, we include \textbf{Diffusion Policy} \cite{chi2023diffusion}, which formulates robot policies as conditional denoising diffusion processes. The 3D baselines comprise voxel-based methods: \textbf{C2F-ARM-BC} \cite{james2022coarse}, which predicts actions in a coarse-to-fine manner via Q-value estimation; \textbf{PerAct} \cite{shridhar2023perceiver}, which detects actions through global self-attention; \textbf{Hiveformer} \cite{guhur2023instruction}, which employs attention across historical features; and \textbf{GNFactor} \cite{ze2023gnfactor}, which co-optimizes a neural scene representation with a PerAct-based policy. We also consider point-based methods: \textbf{PolarNet} \cite{chen2023polarnet}, which computes dense point representations, and multi-view approaches: \textbf{RVT} \cite{goyal2023rvt}, which fuses multi-view predictions via 3D back-projection, and its successor \textbf{RVT2} \cite{goyal2024rvt2}, which enhances precision and efficiency. Furthermore, we include \textbf{Act3D} \cite{gervet2023act3d}, which uses coarse-to-fine 3D featurization (retrained by us for a fair comparison), \textbf{3D Diffuser Actor} \cite{ke20253d}, a 3D diffusion-based policy, and \textbf{ARP} \cite{zhang2025autoregressive}, an autoregressive policy generating hybrid action sequences via a Chunking Causal Transformer.

\subsection*{Implementation Details.}

In the RLBench \textit{Multi-view setup} experiments, during the supervised pre-training phase, our experimental hyperparameter settings comply with APR\cite{zhang2025autoregressive}. In the offline reinforcement learning phase for MVEP, our hyperparameter configurations are detailed in Table \ref{tab:hp-mvep}.

% \begin{table}[ht]
% \centering
% \begin{tabular}{>{\raggedright\arraybackslash}p{0.5\linewidth}>{\raggedright\arraybackslash}p{0.5\linewidth}}
% \toprule
% \textbf{Hyperparameter} & \textbf{Value} \\
% \midrule
% \multicolumn{2}{l}{\textit{MVEP
% }} \\
% $K$ number of the camera viewpoints & 3 \\
% $N_{MVEP}$ The number of input point clouds for MVEP & 2048 \\
% embedding size & 512\\
% \midrule
% % \multicolumn{2}{l}{\textit{sensitive parameters}} \\
% % $\theta$ & $0\sim\pi$ \\
% % $\phi$ & $0\sim2\pi$ \\
% % $r_{min}$ & 0.75 \\
% % $r_{max}$ & 1.3 \\
% % \midrule
% \multicolumn{2}{l}{\textit{Train \& Eval}} \\
% observation(RGBD) & $4 \times 128 \times 128 \times 4$ \\
% re-render image resolution & $224 \times 224$ \\
% maximum evaluation steps & 25 \\
% train epochs & 20 \\
% eval frequency & 100 \\
% batch size & 96 \\
% learning rate & 2.0e-6 \\
% learning rate scheduler & cosine \\
% optimizer & LAMB \\
% \bottomrule
% \end{tabular}
% \vspace{-8pt}
% \caption{Hyperparameters for offline reinforcement learning training in RLBench experiments.}
% \vspace{-10pt}
% \label{tab:hp-mvep}
% \end{table}

\begin{table}[ht]
\centering
\setlength{\tabcolsep}{4pt} % 减少列间距
\begin{tabular}{>{\raggedright\arraybackslash}p{0.45\linewidth}>{\raggedright\arraybackslash}p{0.45\linewidth}}
\toprule
\textbf{Hyperparameter} & \textbf{Value} \\
\midrule
\multicolumn{2}{l}{\textit{MVEP}} \\
$K$ number of the camera viewpoints & 3 \\
$N_{MVEP}$ The number of input point clouds for MVEP & 2048 \\
embedding size & 512\\
\midrule
\multicolumn{2}{l}{\textit{Train \& Eval}} \\
observation(RGBD) & $4 \times 128 \times 128 \times 4$ \\
re-render image resolution & $224 \times 224$ \\
maximum evaluation steps & 25 \\
train epochs & 20 \\
eval frequency & 100 \\
batch size & 96 \\
learning rate & 2.0e-6 \\
learning rate scheduler & cosine \\
optimizer & LAMB \\
\bottomrule
\end{tabular}
\vspace{-8pt}
\caption{Hyperparameters for offline reinforcement learning training in RLBench experiments.}
\vspace{-10pt}
\label{tab:hp-mvep}
\end{table}

\subsection*{Performance-Efficiency Trade-off Analysis.}

\begin{table}[t]
\centering
\scriptsize
\setlength{\tabcolsep}{10pt}
\begin{tabular}{l cc}
\toprule
\textbf{Method}& \textbf{Average Success Rate (\%)}  &\textbf{Average Inference Time(s)}\\
\midrule
ARP& 81.6& 0.394\\
\ModelAbbr& 86.6& 0.436\\\hline
\end{tabular}
\vspace{-10pt}
\caption{Comparison of the average success rate and inference time between ARP and \ModelAbbr\ across 18 tasks.}
\vspace{-15pt}
\label{tab:inference-time-comparison}
\end{table}

During inference, \ModelAbbr\ must predict dynamic multi-view camera poses and continuously adjust the rendering camera pose, resulting in a non-negligible computational overhead. To assess the practicality of the proposed \ModelAbbr\ model, we compare its average task success rate and inference latency with those of the baseline model, ARP. The results, summarized in Table \ref{tab:inference-time-comparison} on the RLBench, show that \ModelAbbr\ achieves a higher average success rate than ARP, while the increase in inference latency remains relatively modest (approximately 10.7\%). These findings suggest that \ModelAbbr\ effectively balances task performance and operational efficiency, with design elements such as sampling acceleration and camera caching playing a key role in enhancing model efficiency.

\subsection*{Data Efficiency Analysis.}

\begin{table}[h]
\centering
\vspace{-12pt}
\resizebox{0.95\linewidth}{!}{%
\small
\setlength{\tabcolsep}{4pt}
\vspace{-10pt}
\begin{tabular}{lcccc}
\toprule
\textbf{Method} & \textbf{20} & \textbf{40} & \textbf{80} & \textbf{100} \\
\midrule
ARP & \(6.7 \pm 2.3\) & \(14.7 \pm 2.3\) & \(28.0 \pm 4.0\) & \(45.3 \pm 6.1\) \\
\ModelAbbr\ (Ours) & \(\mathbf{10.7 \pm 2.3}\) & \(\mathbf{17.3 \pm 2.3}\) & \(\mathbf{30.7 \pm 6.1}\) & \(\mathbf{52.0 \pm 4.0}\) \\
\bottomrule
\end{tabular}
}
\vspace{-10pt}
\caption{Success rate (mean $\pm$ std \%) comparison under varying numbers of demonstrations on the RLBench \textit{Put in Cupboard} task.}
\vspace{-10pt}
\label{tab:demo_ablation}
\end{table}

We conduct a demonstration ablation study on the RLBench \textit{Put in Cupboard} task. As shown in Table~\ref{tab:demo_ablation}, \ModelAbbr\ consistently outperforms the baseline ARP across all demonstration sizes (20, 40, 80, and 100). Notably, the performance gap widens as the number of demonstrations increases, with \ModelAbbr\ achieving a success rate of \(\mathbf{52.0 \pm 4.0\%}\) compared to \(45.3 \pm 6.1\%\) for ARP at 100 demonstrations. This trend demonstrates that while more data benefits both methods, the superior sample efficiency and final performance of \ModelAbbr\ are inherently due to our proposed architectural improvements, not simply the result of scaling the dataset.

\subsection*{Results on RLBench-OG.}

We conducted an evaluation of \ModelAbbr, RVT2, ARP, and Diffusion Policy on the RLBench-OG benchmark. Detailed performance across 10 tasks under 8 variations is presented in Tables~\ref{tab:tvve-on-og}, \ref{tab:rvt2-on-og}, \ref{tab:arp-on-og}, and~\ref{tab:dp-on-og}. The reported success rates correspond to the mean $\pm$ standard deviation from three independent trials.

\begin{table*}[htbp]
\centering
\scriptsize
\resizebox{1.0\linewidth}{!}{%
\small
\setlength{\tabcolsep}{2.0pt}
\begin{tabular}{lccccccccc}
\toprule
 \textbf{Task Name} & \textbf{Occlusion 1} & \textbf{Occlusion 2} & \textbf{Light Color} & \textbf{Table Color} & \textbf{Table Texture} & \textbf{Distractor} & \textbf{Background Texture} & \textbf{Camera Pose} & \textbf{Variant Mean} \\
\midrule
basketball\_in\_hoop & $ 100.0\pm0.0 $ & $ 100.0\pm0.0 $ & $ 100.0\pm0.0 $ & $ 92.0\pm0.0 $ & $ 96.0\pm0.0 $ & $ 70.0\pm2.0 $ & $ 100.0\pm0.0 $ & $ 100.0\pm0.0 $ & $ 94.8\pm9.8 $ \\
scoop\_with\_spatula & $ 88.0\pm8.0 $ & $ 88.0\pm5.7 $ & $ 64.0\pm0.0 $ & $ 76.0\pm0.0 $ & $ 86.0\pm2.0 $ & $ 68.0\pm4.0 $ & $ 92.0\pm4.0 $ & $ 89.3\pm3.8 $ & $ 81.4\pm10.8 $ \\
take\_plate\_off\_colored\_dish\_rack & $ 100.0\pm0.0 $ & $ 98.7\pm1.9 $ & $ 76.0\pm4.0 $ & $ 94.0\pm6.0 $ & $ 92.0\pm0.0 $ & $ 100.0\pm0.0 $ & $ 98.7\pm1.9 $ & $ 100.0\pm0.0 $ & $ 94.9\pm8.2 $ \\
water\_plants & $ 88.0\pm4.0 $ & $ 26.7\pm3.8 $ & $ 16.0\pm4.0 $ & $ 16.0\pm0.0 $ & $ 20.0\pm4.0 $ & $ 22.7\pm1.9 $ & $ 24.0\pm4.0 $ & $ 24.0\pm3.3 $ & $ 29.7\pm22.6 $ \\
block\_pyramid & $ 8.0\pm8.0 $ & $ 0.0\pm0.0 $ & $ 22.0\pm6.0 $ & $ 8.0\pm0.0 $ & $ 8.0\pm0.0 $ & $ 12.0\pm0.0 $ & $ 16.0\pm4.0 $ & $ 18.0\pm6.0 $ & $ 11.5\pm7.9 $ \\
solve\_puzzle & $ 2.0\pm2.0 $ & $ 0.0\pm0.0 $ & $ 14.0\pm6.0 $ & $ 14.0\pm2.0 $ & $ 10.0\pm2.0 $ & $ 12.0\pm0.0 $ & $ 24.0\pm0.0 $ & $ 22.0\pm2.0 $ & $ 12.2\pm8.3 $ \\
take\_usb\_out\_of\_computer & $ 100.0\pm0.0 $ & $ 100.0\pm0.0 $ & $ 97.3\pm1.9 $ & $ 100.0\pm0.0 $ & $ 96.0\pm0.0 $ & $ 82.7\pm1.9 $ & $ 100.0\pm0.0 $ & $ 100.0\pm0.0 $ & $ 97.0\pm5.7 $ \\
close\_drawer & $ 100.0\pm0.0 $ & $ 100.0\pm0.0 $ & $ 92.0\pm0.0 $ & $ 96.0\pm0.0 $ & $ 96.0\pm0.0 $ & $ 98.7\pm1.9 $ & $ 98.7\pm1.9 $ & $ 98.7\pm1.9 $ & $ 97.5\pm2.8 $ \\
straighten\_rope & $ 64.0\pm0.0 $ & $ 0.0\pm0.0 $ & $ 58.0\pm2.0 $ & $ 50.0\pm2.0 $ & $ 64.0\pm8.0 $ & $ 36.0\pm0.0 $ & $ 89.3\pm5.0 $ & $ 80.0\pm4.0 $ & $ 55.2\pm26.2 $ \\
toilet\_seat\_down & $ 100.0\pm0.0 $ & $ 66.7\pm8.2 $ & $ 97.3\pm1.9 $ & $ 100.0\pm0.0 $ & $ 100.0\pm0.0 $ & $ 100.0\pm0.0 $ & $ 100.0\pm0.0 $ & $ 100.0\pm0.0 $ & $ 95.5\pm11.3 $ \\
\midrule
\textbf{Task Mean} & $ 75.0\pm1.2 $ & $ 58.0\pm1.1 $ & $ 63.7\pm1.1 $ & $ 64.6\pm0.7 $ & $ 66.8\pm0.9 $ & $ 60.2\pm0.6 $ & $ 74.3\pm0.9 $ & $ 73.2\pm0.9 $ & $ 67.0\pm6.2 $ \\
\bottomrule
\end{tabular}
}
\vspace{-10pt}
\caption{Success Rates of \ModelAbbr\ under Different Perturbations of RLBench-OG.}
\vspace{-10pt}
\label{tab:tvve-on-og}
\end{table*}

\begin{table*}[htbp]
\centering
\scriptsize
\resizebox{1.0\linewidth}{!}{%
\small
\setlength{\tabcolsep}{2.0pt}
\begin{tabular}{lccccccccc}
\toprule
 \textbf{Task Name} & \textbf{Occlusion 1} & \textbf{Occlusion 2} & \textbf{Light Color} & \textbf{Table Color} & \textbf{Table Texture} & \textbf{Distractor} & \textbf{Background Texture} & \textbf{Camera Pose} & \textbf{Variant Mean} \\
\midrule
basketball\_in\_hoop & $ 100.0\pm0.0 $ & $ 100.0\pm0.0 $ & $ 88.8\pm1.6 $ & $ 96.0\pm0.0 $ & $ 99.2\pm1.6 $ & $ 88.0\pm0.0 $ & $ 100.0\pm0.0 $ & $ 100.0\pm0.0 $ & $ 96.5\pm4.8 $ \\
scoop\_with\_spatula & $ 82.7\pm7.5 $ & $ 84.0\pm6.5 $ & $ 75.4\pm2.6 $ & $ 84.0\pm5.7 $ & $ 88.0\pm5.7 $ & $ 76.0\pm4.0 $ & $ 98.0\pm2.0 $ & $ 96.0\pm0.0 $ & $ 85.5\pm7.7 $ \\
take\_plate\_off\_colored\_dish\_rack & $ 98.7\pm1.9 $ & $ 65.3\pm6.8 $ & $ 60.0\pm4.6 $ & $ 88.8\pm3.9 $ & $ 84.8\pm4.7 $ & $ 100.0\pm0.0 $ & $ 100.0\pm0.0 $ & $ 100.0\pm0.0 $ & $ 87.2\pm15.2 $ \\
water\_plants & $ 72.0\pm8.6 $ & $ 20.0\pm5.7 $ & $ 12.0\pm5.1 $ & $ 9.6\pm9.3 $ & $ 23.0\pm5.9 $ & $ 22.0\pm2.0 $ & $ 26.0\pm2.0 $ & $ 32.0\pm0.0 $ & $ 27.1\pm18.3 $ \\
block\_pyramid & $ 4.0\pm3.3 $ & $ 0.0\pm0.0 $ & $ 0.0\pm0.0 $ & $ 0.0\pm0.0 $ & $ 0.0\pm0.0 $ & $ 0.0\pm0.0 $ & $ 0.0\pm0.0 $ & $ 0.0\pm0.0 $ & $ 0.5\pm1.3 $ \\
solve\_puzzle & $ 0.0\pm0.0 $ & $ 0.0\pm0.0 $ & $ 9.3\pm3.0 $ & $ 8.0\pm3.6 $ & $ 12.0\pm5.1 $ & $ 20.0\pm0.0 $ & $ 16.0\pm3.3 $ & $ 25.3\pm3.8 $ & $ 11.3\pm8.4 $ \\
take\_usb\_out\_of\_computer & $ 100.0\pm0.0 $ & $ 100.0\pm0.0 $ & $ 100.0\pm0.0 $ & $ 100.0\pm0.0 $ & $ 95.2\pm1.6 $ & $ 98.0\pm2.0 $ & $ 98.7\pm1.9 $ & $ 100.0\pm0.0 $ & $ 99.0\pm1.6 $ \\
close\_drawer & $ 100.0\pm0.0 $ & $ 100.0\pm0.0 $ & $ 99.3\pm1.5 $ & $ 97.6\pm2.0 $ & $ 98.4\pm2.0 $ & $ 96.0\pm0.0 $ & $ 100.0\pm0.0 $ & $ 100.0\pm0.0 $ & $ 98.9\pm1.4 $ \\
straighten\_rope & $ 70.7\pm10.0 $ & $ 0.0\pm0.0 $ & $ 56.7\pm5.8 $ & $ 37.6\pm5.4 $ & $ 44.0\pm5.1 $ & $ 36.0\pm0.0 $ & $ 86.7\pm5.0 $ & $ 88.0\pm4.0 $ & $ 52.5\pm27.6 $ \\
toilet\_seat\_down & $ 100.0\pm0.0 $ & $ 0.0\pm0.0 $ & $ 100.0\pm0.0 $ & $ 100.0\pm0.0 $ & $ 100.0\pm0.0 $ & $ 98.0\pm2.0 $ & $ 100.0\pm0.0 $ & $ 100.0\pm0.0 $ & $ 87.2\pm33.0 $ \\
\midrule
\textbf{Task Mean} & $ 72.8\pm2.0 $ & $ 46.9\pm0.4 $ & $ 60.8\pm1.2 $ & $ 61.8\pm1.1 $ & $ 64.0\pm2.0 $ & $ 63.4\pm0.2 $ & $ 72.6\pm0.6 $ & $ 74.0\pm0.8 $ & $ 64.5\pm8.3 $ \\
\bottomrule
\end{tabular}
}
\vspace{-10pt}
\caption{Success Rates of RVT2 under Different Perturbations of RLBench-OG.}
\vspace{-10pt}
\label{tab:rvt2-on-og}
\end{table*}

\begin{table*}[htbp]
\centering
\scriptsize
\resizebox{1.0\linewidth}{!}{%
\small
\setlength{\tabcolsep}{2.0pt}
\begin{tabular}{lccccccccc}
\toprule
 \textbf{Task Name} & \textbf{Occlusion 1} & \textbf{Occlusion 2} & \textbf{Light Color} & \textbf{Table Color} & \textbf{Table Texture} & \textbf{Distractor} & \textbf{Background Texture} & \textbf{Camera Pose} & \textbf{Variant Mean} \\
\midrule
basketball\_in\_hoop & $ 100.0\pm0.0 $ & $ 100.0\pm0.0 $ & $ 86.0\pm2.0 $ & $ 76.0\pm0.0 $ & $ 86.0\pm2.0 $ & $ 84.0\pm0.0 $ & $ 100.0\pm0.0 $ & $ 100.0\pm0.0 $ & $ 91.5\pm9.0 $ \\
scoop\_with\_spatula & $ 88.0\pm4.0 $ & $ 76.0\pm8.0 $ & $ 62.0\pm2.0 $ & $ 78.0\pm2.0 $ & $ 86.0\pm2.0 $ & $ 82.0\pm2.0 $ & $ 86.7\pm10.0 $ & $ 84.0\pm8.0 $ & $ 80.3\pm9.8 $ \\
take\_plate\_off\_colored\_dish\_rack & $ 100.0\pm0.0 $ & $ 76.0\pm4.0 $ & $ 74.0\pm6.0 $ & $ 96.0\pm0.0 $ & $ 94.0\pm6.0 $ & $ 100.0\pm0.0 $ & $ 100.0\pm0.0 $ & $ 100.0\pm0.0 $ & $ 92.5\pm10.9 $ \\
water\_plants & $ 66.0\pm14.0 $ & $ 10.0\pm2.0 $ & $ 10.0\pm2.0 $ & $ 12.0\pm0.0 $ & $ 10.7\pm3.8 $ & $ 13.3\pm7.5 $ & $ 16.0\pm8.0 $ & $ 21.3\pm3.8 $ & $ 19.9\pm19.0 $ \\
block\_pyramid & $ 14.0\pm2.0 $ & $ 0.0\pm0.0 $ & $ 1.0\pm1.7 $ & $ 5.3\pm3.8 $ & $ 0.8\pm1.6 $ & $ 2.4\pm3.2 $ & $ 2.0\pm3.5 $ & $ 6.4\pm8.2 $ & $ 4.0\pm5.7 $ \\
solve\_puzzle & $ 0.0\pm0.0 $ & $ 0.0\pm0.0 $ & $ 14.0\pm2.0 $ & $ 8.0\pm0.0 $ & $ 6.0\pm2.0 $ & $ 16.0\pm8.6 $ & $ 8.0\pm4.0 $ & $ 14.0\pm6.0 $ & $ 8.2\pm7.1 $ \\
take\_usb\_out\_of\_computer & $ 100.0\pm0.0 $ & $ 100.0\pm0.0 $ & $ 100.0\pm0.0 $ & $ 98.0\pm2.0 $ & $ 96.0\pm0.0 $ & $ 94.0\pm2.0 $ & $ 100.0\pm0.0 $ & $ 98.7\pm1.9 $ & $ 98.3\pm2.4 $ \\
close\_drawer & $ 100.0\pm0.0 $ & $ 100.0\pm0.0 $ & $ 96.0\pm0.0 $ & $ 98.0\pm2.0 $ & $ 100.0\pm0.0 $ & $ 94.0\pm2.0 $ & $ 100.0\pm0.0 $ & $ 100.0\pm0.0 $ & $ 98.5\pm2.4 $ \\
straighten\_rope & $ 62.0\pm6.0 $ & $ 4.0\pm0.0 $ & $ 60.0\pm0.0 $ & $ 60.0\pm0.0 $ & $ 46.0\pm6.0 $ & $ 42.0\pm6.0 $ & $ 68.0\pm8.0 $ & $ 74.0\pm2.0 $ & $ 52.0\pm21.2 $ \\
toilet\_seat\_down & $ 100.0\pm0.0 $ & $ 60.0\pm4.0 $ & $ 94.7\pm1.9 $ & $ 96.0\pm0.0 $ & $ 88.0\pm0.0 $ & $ 96.0\pm0.0 $ & $ 100.0\pm0.0 $ & $ 98.7\pm1.9 $ & $ 91.7\pm12.6 $ \\
\midrule
\textbf{Task Mean} & $ 73.0\pm1.6 $ & $ 52.6\pm1.0 $ & $ 59.8\pm0.8 $ & $ 62.7\pm0.5 $ & $ 61.3\pm1.0 $ & $ 62.4\pm1.4 $ & $ 68.1\pm1.6 $ & $ 69.7\pm1.4 $ & $ 63.7\pm6.1 $ \\
\bottomrule
\end{tabular}
}
\vspace{-10pt}
\caption{Success Rates of ARP under Different Perturbations of RLBench-OG.}
\vspace{-10pt}
\label{tab:arp-on-og}
\end{table*}

\begin{table*}[htbp]
\centering
\scriptsize
\resizebox{1.0\linewidth}{!}{%
\small
\setlength{\tabcolsep}{2.0pt}
\begin{tabular}{lccccccccc}
\toprule
 \textbf{Task Name} & \textbf{Occlusion 1} & \textbf{Occlusion 2} & \textbf{Light Color} & \textbf{Table Color} & \textbf{Table Texture} & \textbf{Distractor} & \textbf{Background Texture} & \textbf{Camera Pose} & \textbf{Variant Mean} \\
\midrule
basketball\_in\_hoop & $ 0.0\pm0.0 $ & $ 0.0\pm0.0 $ & $ 0.0\pm0.0 $ & $ 0.0\pm0.0 $ & $ 0.0\pm0.0 $ & $ 0.0\pm0.0 $ & $ 0.0\pm0.0 $ & $ 0.0\pm0.0 $ & $ 0.0\pm0.0 $ \\
scoop\_with\_spatula & $ 0.0\pm0.0 $ & $ 0.0\pm0.0 $ & $ 0.0\pm0.0 $ & $ 0.0\pm0.0 $ & $ 0.0\pm0.0 $ & $ 1.3\pm1.9 $ & $ 1.3\pm1.9 $ & $ 0.0\pm0.0 $ & $ 0.3\pm1.1 $ \\
take\_plate\_off\_colored\_dish\_rack & $ 4.0\pm0.0 $ & $ 0.0\pm0.0 $ & $ 0.0\pm0.0 $ & $ 0.0\pm0.0 $ & $ 2.0\pm2.0 $ & $ 0.0\pm0.0 $ & $ 0.0\pm0.0 $ & $ 0.0\pm0.0 $ & $ 0.8\pm1.6 $ \\
water\_plants & $ 2.0\pm2.0 $ & $ 4.0\pm0.0 $ & $ 8.0\pm0.0 $ & $ 4.0\pm3.3 $ & $ 2.0\pm2.0 $ & $ 8.0\pm3.3 $ & $ 2.0\pm2.0 $ & $ 2.0\pm2.0 $ & $ 4.0\pm3.3 $ \\
block\_pyramid & $ 0.0\pm0.0 $ & $ 0.0\pm0.0 $ & $ 0.0\pm0.0 $ & $ 0.0\pm0.0 $ & $ 0.0\pm0.0 $ & $ 0.0\pm0.0 $ & $ 0.0\pm0.0 $ & $ 0.0\pm0.0 $ & $ 0.0\pm0.0 $ \\
solve\_puzzle & $ 0.0\pm0.0 $ & $ 0.0\pm0.0 $ & $ 0.0\pm0.0 $ & $ 0.0\pm0.0 $ & $ 0.0\pm0.0 $ & $ 0.0\pm0.0 $ & $ 0.0\pm0.0 $ & $ 0.0\pm0.0 $ & $ 0.0\pm0.0 $ \\
take\_usb\_out\_of\_computer & $ 92.0\pm8.0 $ & $ 82.0\pm2.0 $ & $ 61.3\pm3.8 $ & $ 58.7\pm5.0 $ & $ 66.0\pm2.0 $ & $ 76.0\pm0.0 $ & $ 65.3\pm7.5 $ & $ 64.0\pm12.0 $ & $ 70.7\pm12.5 $ \\
close\_drawer & $ 76.0\pm0.0 $ & $ 56.0\pm8.0 $ & $ 86.7\pm5.0 $ & $ 88.0\pm3.3 $ & $ 72.0\pm0.0 $ & $ 84.0\pm3.3 $ & $ 92.0\pm3.3 $ & $ 74.0\pm2.0 $ & $ 78.6\pm11.6 $ \\
straighten\_rope & $ 0.0\pm0.0 $ & $ 0.0\pm0.0 $ & $ 0.0\pm0.0 $ & $ 0.0\pm0.0 $ & $ 0.0\pm0.0 $ & $ 0.0\pm0.0 $ & $ 0.0\pm0.0 $ & $ 0.0\pm0.0 $ & $ 0.0\pm0.0 $ \\
toilet\_seat\_down & $ 100.0\pm0.0 $ & $ 92.0\pm8.0 $ & $ 73.3\pm5.0 $ & $ 74.7\pm1.9 $ & $ 84.0\pm4.0 $ & $ 74.7\pm13.2 $ & $ 88.0\pm8.6 $ & $ 82.0\pm2.0 $ & $ 83.6\pm11.1 $ \\
\midrule
\textbf{Task Mean} & $ 27.4\pm0.8 $ & $ 23.4\pm1.1 $ & $ 22.9\pm0.8 $ & $ 22.5\pm0.7 $ & $ 22.6\pm0.5 $ & $ 24.4\pm1.4 $ & $ 24.9\pm1.2 $ & $ 22.2\pm1.2 $ & $ 23.8\pm1.9 $ \\
\bottomrule
\end{tabular}
}
\vspace{-10pt}
\caption{Success Rates of Diffusion Policy under Different Perturbations of RLBench-OG.}
\vspace{-10pt}
\label{tab:dp-on-og}
\end{table*}

% \subsection*{Results on Colosseum.}

\subsection*{Experimental Results and Analysis on TaskMoE.}

\begin{table*}[t]
\centering
\scriptsize
\resizebox{1.005\linewidth}{!}{%
\small
\setlength{\tabcolsep}{2.0pt}
\begin{tabular}{l|cccccccccccccccccc|ccc}
\toprule
\multicolumn{19}{c|}{\textbf{Seen}} & \multicolumn{3}{c}{\textbf{Unseen}} \\
\cmidrule(lr){1-19} \cmidrule(lr){20-22}
\makecell{Avg. SR\\(\%)} $\uparrow$ & \makecell{Insert onto\\Square Peg} & \makecell{Put Item\\in Drawer} & \makecell{Reach and\\Drag} & \makecell{Turn Tap} & \makecell{Slide Block to\\Color Target} & \makecell{Open Drawer} & \makecell{Put Groceries\\in Cupboard} & \makecell{Place Shape in\\Shape Sorter} & \makecell{Put Money\\in Safe} & \makecell{Push Buttons} & \makecell{Close Jar} & \makecell{Stack Blocks} & \makecell{Place Cups} & \makecell{Place Wine at\\Rack Location} & \makecell{Light Bulb\\In} & \makecell{Sweep to\\Dustpan of Size} & \makecell{Meat off\\Grill} & \makecell{Stack Cups} & \makecell{Water Plants} & \makecell{Close Drawer} & \makecell{Toilet Seat\\Down} \\
\midrule
80.2 & 24.0 & 100.0 & 96.0 & 100.0 & 100.0 & 88.0 & 68.0 & 32.0 & 96.0 & 96.0 & 100.0 & 68.0 & 36.0 & 96.0 & 88.0 & 96.0 & 96.0 & 64.0 & 12.0 & 44.0 & 100.0 \\
\bottomrule
\end{tabular}
}
\vspace{-8pt}
\caption{\textbf{Performance with $N_G=8,N_E=16$ TaskMoE in RLBench.} TaskMoE enhances generalization to unseen tasks.}
\vspace{-15pt}
\label{tab:taskmoe-gen-unseen}
\end{table*}

\begin{figure}[t]
    \centering
    \vspace{-5pt}
\includegraphics[width=1\linewidth]{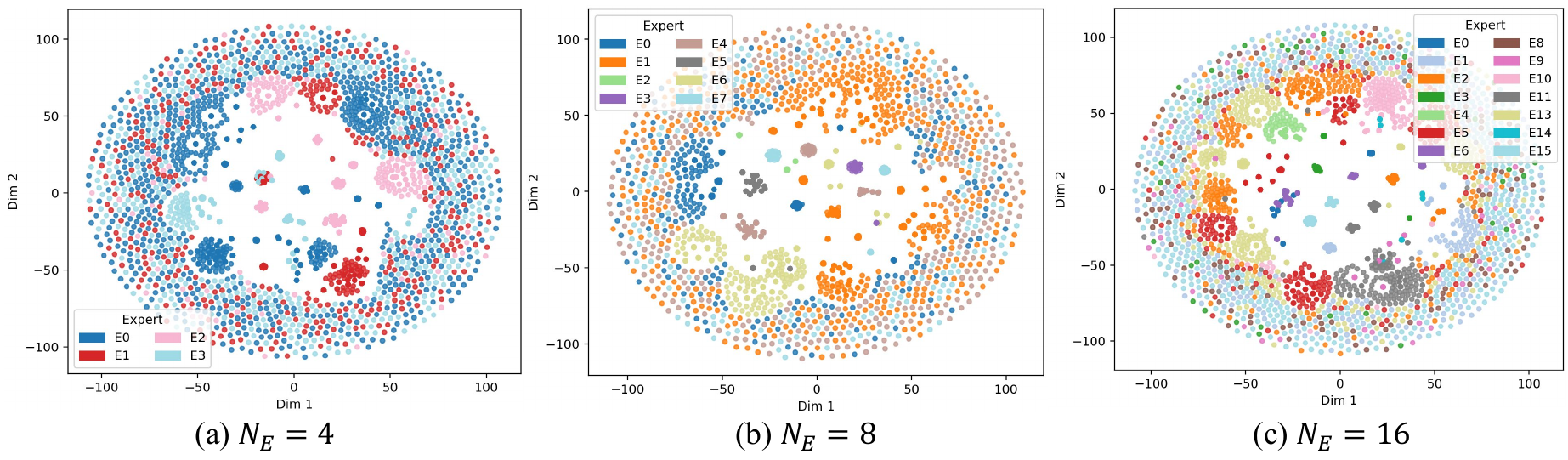}
\vspace{-20pt}
    \caption{t-SNE visualization of instruction embeddings routed through the 4-expert (a), 8-expert (b) and 16-expert (c) TaskMoE. In the 4-expert TaskMoE (a), each expert bears a high workload, their specialization is limited, and it is difficult to achieve fine-grained partitioning. In the 8-expert TaskMoE (b), inner-region experts (E0, E1, E2, E3, E4, E5, E6, E7) form compact, semantically coherent clusters, while outer-ring embeddings are dominated by E1, E4 and E7, indicating fallback routing for diverse or weakly structured instructions. Compared to the 8-expert configuration, the 16-expert TaskMoE exhibits more balanced expert utilization. $N_E$ denotes the number of experts.}
\vspace{-15pt}
    \label{fig:T-SNE-on-taskmoe}
\end{figure}

In the RLBench \textit{Multi-view setup} experiments, we conducted individual ablation studies on the components of TaskMoE, and visualized and analyzed the routing behavior of the gate-expert at both the \textbf{instruction-level} and \textbf{task-level}.

\paragraph{Instruction-level Routing Behavior Analysis.} We compare the routing behavior of the 16-expert Task-MoE with the 4-expert and 8-expert configuration using instruction-level t-SNE embeddings, as shown in Fig.~\ref{fig:T-SNE-on-taskmoe}. The 16-expert model exhibits markedly improved semantic disentanglement: the embedding manifold contains a substantially larger number of compact and well-separated clusters, each corresponding to an expert with strong specialization. Expert utilization becomes significantly more balanced, with reduced reliance on fallback experts for long-tail or weakly structured instructions. The outer-ring region—typically occupied by generic or ambiguous instructions—is shared by a diverse subset of experts rather than collapsing onto one or two dominant experts. This indicates that the expanded expert capacity enables finer-grained semantic partitioning and mitigates routing collapse. In contrast, the 8-expert model shows partially collapsed routing behavior. Although several inner-region experts still form coherent semantic clusters, the outer ring is dominated by E1, E4 and E7, which assume the role of catch-all experts. Their dispersed and high-density presence suggests that the gate struggles to allocate long-tail and noisy instructions across multiple experts. Consequently, the semantic boundaries between experts remain coarse, limiting the degree of modularization and reducing the benefits of MoE specialization. Overall, the comparison demonstrates that increasing the number of experts from 8 to 16 leads to more stable routing, finer semantic granularity, and healthier expert balancing, thereby enabling a more expressive and disentangled mixture-of-experts structure.

\begin{figure}[t]
    \centering
    \vspace{-5pt}
\includegraphics[width=1\linewidth]{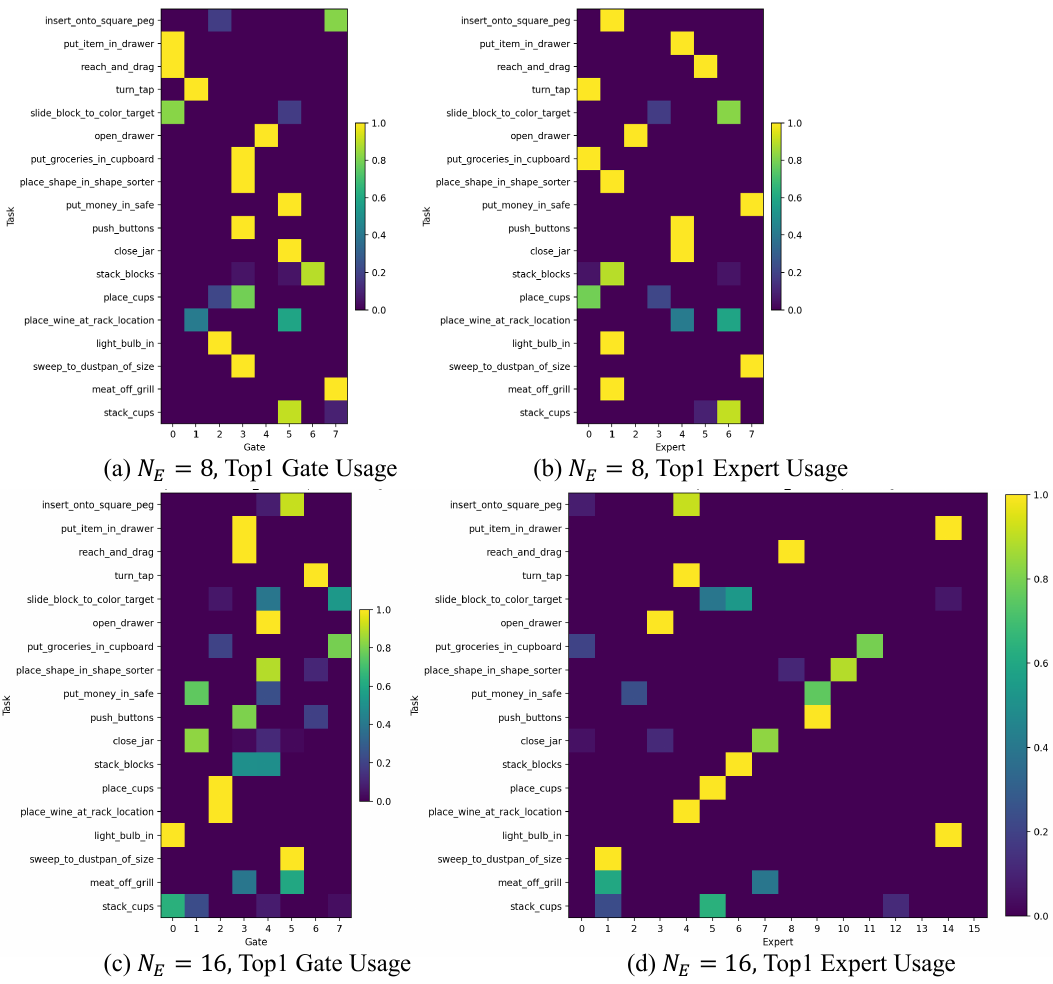}
\vspace{-15pt}
    \caption{Task-wise gate and expert usage visualization for the proposed MoE layer. Both 8-expert (a,b) and 16-expert (c,d) configurations demonstrate clear task-dependent routing, while larger expert capacity allows finer-grained specialization without leading to expert collapse.}
\vspace{-15pt}
    \label{fig:taskmoe8vs16}
\end{figure}

\paragraph{Task-level Routing Behavior Analysis.} 
We analyze the gate and expert routing behavior by visualizing the task-wise distributions. Such visualization allows us to examine whether different RLBench tasks activate distinct subsets of gates and experts, revealing the degree of task specialization, routing stability, and expert utilization diversity within the TaskMoE architecture.

Fig.~\ref{fig:taskmoe8vs16} shows per–task top-1 gate/expert usage for $N_E\!=\!8$ (a,b) and $N_E\!=\!16$ (c,d). With 8 experts, every task settles on 1–2 gates (e.g., \textit{place\_shape\_in\_shape\_sorter} and \textit{put\_groceries\_in\_cupboard} both enter G3; \textit{place\_cups} enters G2/G3), and all gates/experts are exercised, confirming the task-aware gate design learns a stable task$\rightarrow$gate partition without idle capacity. Increasing to 16 experts keeps the gate partition similar but further splits the downstream experts: tasks that shared a single expert at $N_E\!=\!8$ now route to distinct experts (e.g., \textit{meat\_off\_grill} disperses over E1/E7 (from E1), \textit{place\_shape\_in\_shape\_sorter} shifts to E8/E10 (from E1), \textit{put\_money\_in\_safe} to E2/E9 (from E7), yielding finer skill specialization. This validates that the two-stage gate$\rightarrow$expert routing, combined with entropy/semantic regularization and a larger $N_E$, not only provides additional capacity that is fully utilized without collapsing to a few experts but also enhances sparsity, which optimizes computational efficiency and improves generalization by distributing representations more distinctly.

Moreover, Fig.~\ref{fig:taskmoe8vs16}(c,d) shows the two-stage router differentiates fine-grained skills even when tasks share the same gate. For example, the semantically related tasks \textit{place\_cups} and \textit{place\_wine\_at\_rack\_location} both enter the same gate (G2) at $N_E\!=\!16$, yet their traffic is split to different experts downstream (e.g., \textit{place\_cups} to E5 vs. \textit{place\_wine\_at\_rack\_location} to E4). This demonstrates that the gate layer captures coarse task affinity while the expert layer refines the routing to task-specific specialists, confirming the effectiveness of the two-stage gate$\rightarrow$expert design in achieving fine-grained routing without losing shared structure.

\paragraph{Generalization.}

We conduct experiments in RLBench, training on tasks such as \textit{Insert\_Peg} and \textit{Put\_Item\_Drawer}, and others, and is subsequently evaluated on a set of unseen tasks. Our model achieves high success rates in completing certain tasks not encountered during training, demonstrating the generalization capability of our architecture. The results are shown in Table~\ref{tab:taskmoe-gen-unseen}.

\section*{Appendix D. Implementation details and information about the RLBench and RLBench-OG}

\subsection*{RLBench.}

To evaluate the effectiveness and generalization capability of the proposed \ModelAbbr\ for multi-task robotic manipulation, we conduct experiments on 18 diverse tasks from the RLBench benchmark, including \textit{close jar}, \textit{stack blocks}, and so on. \ModelAbbr\ is trained with 100 demonstrations per task. Example demonstration samples for each task are shown in Fig.~\ref{fig:app-rlbench}. During evaluation, the model is tested with 25 demonstrations using the same settings. For a more intuitive understanding, we also include corresponding video results.

\subsection*{RLBench-OG: Task Definition and Visualization.}

\begin{figure*}[h!]
    \centering
    \vspace{-5pt}
    \includegraphics[width=0.85\linewidth]{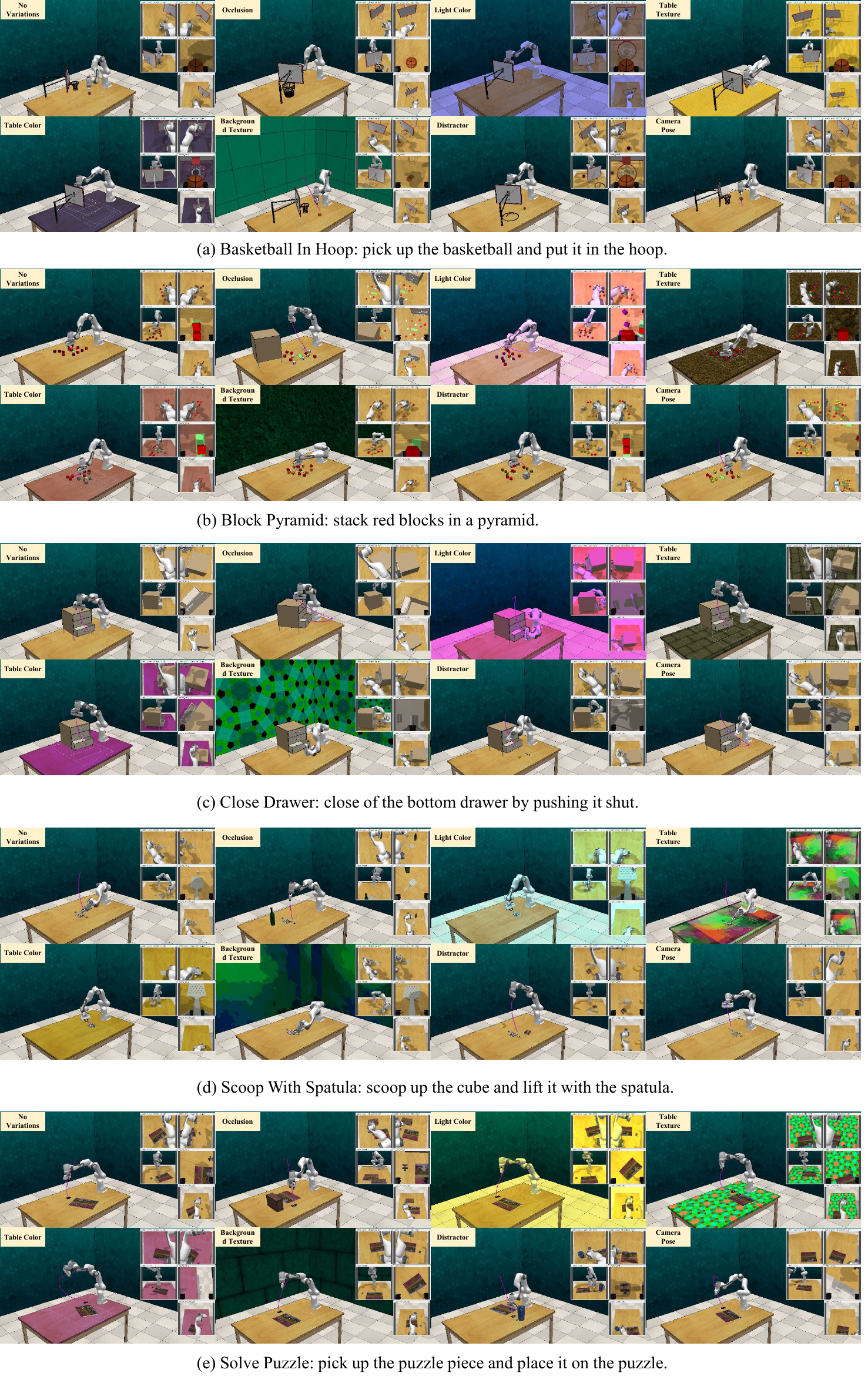}
    \vspace{-10pt}
    \caption{Visualization of different variants for the \textbf{basketball\_in\_hoop}, \textbf{block\_pyramid}, \textbf{close\_drawer}, \textbf{scoop\_with\_spatula}, \textbf{solve\_puzzle} tasks.}
    % \vspace{-15pt}
    \label{fig:og-1}
\end{figure*}

\begin{figure*}[h!]
    \centering
    \vspace{-5pt}
    \includegraphics[width=0.85\linewidth]{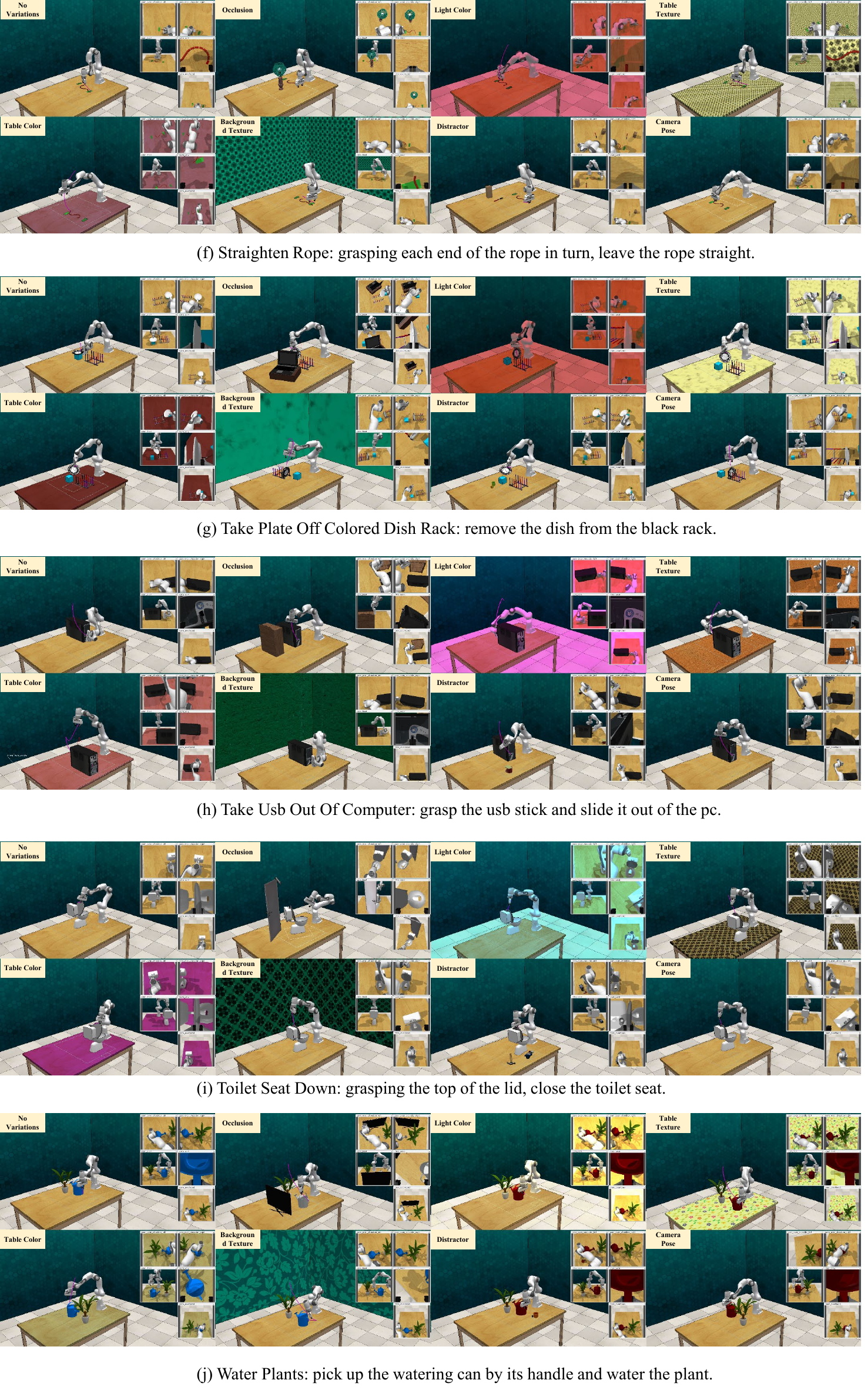}
    \vspace{-10pt}
    \caption{Visualization of different variants for the \textbf{straighten\_rope}, \textbf{take\_plate\_off\_colored\_dish\_rack}, \textbf{take\_usb\_out\_of\_computer}, \textbf{toilet\_seat\_down}, \textbf{water\_plants} tasks.}
    % \vspace{-15pt}
    \label{fig:og-2}
\end{figure*}

RLBench-OG is derived from the RLBench benchmark and is designed to evaluate the robustness of models under occlusion as well as their generalization capability under various environmental perturbations. RLBench-OG selects ten tasks from the original RLBench task list, covering both simple scenarios (e.g., \textit{take\_usb\_out\_of\_computer}) and more complex long-horizon tasks (e.g., \textit{block\_pyramid}). The benchmark consists of two components: an \textbf{Occlusion Suite} and a \textbf{Generalization Suite}. We detail both components below. For the visualization of different variant settings corresponding to each task, see Fig. ~\ref{fig:og-1} and Fig. ~\ref{fig:og-2}.

\subsubsection*{Occlusion Suite}
Occlusion refers to situations in which the line of sight of the camera to key task-relevant regions is fully or partially blocked, leading to incomplete observations and degraded performance of state estimation and action execution. In the occlusion suite, we introduce occlusions to the \textit{front\_camera} through two mechanisms:

\begin{enumerate}
    \item \textbf{Self-occlusion by object pose perturbation.} We modify the position or orientation of task-relevant objects such that essential interaction points become occluded. These occluded regions are often critical for task completion, such as the drawer handle in the \textit{close\_drawer} task.
    \item \textbf{Occlusion by external distractors.} We place task-irrelevant objects---such as cabinets, TVs, or doors---in front of the workspace to partially block the scene, leading to incomplete visibility of key regions.
\end{enumerate}

\paragraph{Task Construction.}
The following describes how occlusions are introduced for each of the ten tasks:

\begin{itemize}
    \item \textbf{basketball\_in\_hoop:} Basket and trash can poses are perturbed to occlude the basketball.
    \item \textbf{block\_pyramid:} A cabinet is placed in front of the workspace to occlude part of the blocks.
    \item \textbf{close\_drawer:} The drawer is rotated such that its geometry occludes the handle.
    \item \textbf{scoop\_with\_spatula:} A wine bottle is positioned to block the target cube.
    \item \textbf{solve\_puzzle:} A storage cabinet is placed to occlude puzzle pieces.
    \item \textbf{straighten\_rope:} A desk lamp is placed in front of one end of the rope.
    \item \textbf{take\_plate\_off\_colored\_dish\_rack:} A box with a laptop blocks visibility of the plate.
    \item \textbf{take\_usb\_out\_of\_computer:} A cabinet blocks the USB port area.
    \item \textbf{toilet\_seat\_down:} A door is placed such that it occludes the toilet seat.
    \item \textbf{water\_plants:} A television partially blocks both the watering can and the plant.
\end{itemize}

\paragraph{Experimental Settings.}
We evaluate models under two occlusion levels, \textbf{Occlusion~1} and \textbf{Occlusion~2}. For \textbf{Occlusion~1}, models are both trained and tested directly under the occluded task configurations; for \textbf{Occlusion~2}, models are trained in the original RLBench task settings and then evaluated in a zero-shot manner under occluded conditions.

\subsubsection*{Generalization Suite}

The Generalization Suite evaluates robustness to environment-conditioned variations. Based on the same ten tasks, we construct six types of environment variations, each modifying exactly one factor while keeping all others unchanged. Following the pipeline from the COLOSSEUM~\cite{pumacay2024colosseum}, we specify variation types using \textit{yaml} configuration files and data collection procedures via \textit{json} metadata.

\paragraph{Variation Types.}
\begin{itemize}
    \item \textbf{light\_color:} RGB values are sampled within predefined ranges and applied to directional lights.
    \item \textbf{table\_texture:} A texture is sampled from a texture dataset and applied to the table.
    \item \textbf{table\_color:} RGB values are sampled within predefined ranges and applied to the table surface.
    \item \textbf{background\_texture:} A background texture is randomly sampled and applied.
    \item \textbf{distractor:} Two distractor objects are sampled from a 3D asset dataset and spawned within the workspace boundary.
    \item \textbf{camera\_pose:} Camera position and orientation offsets are sampled and applied to the front, left-shoulder, and right-shoulder cameras.
\end{itemize}

\paragraph{Experimental Settings.}
For the Generalization Suite, models are trained in the original RLBench environment and subsequently evaluated in a zero-shot manner across various generalization variants. For all variants, we collect 25 validation episodes for each task.

\section*{Appendix E. Multi-view Re-rendering Visualization Results}

\begin{figure*}[t]
    \centering
    \includegraphics[width=0.7\textwidth]{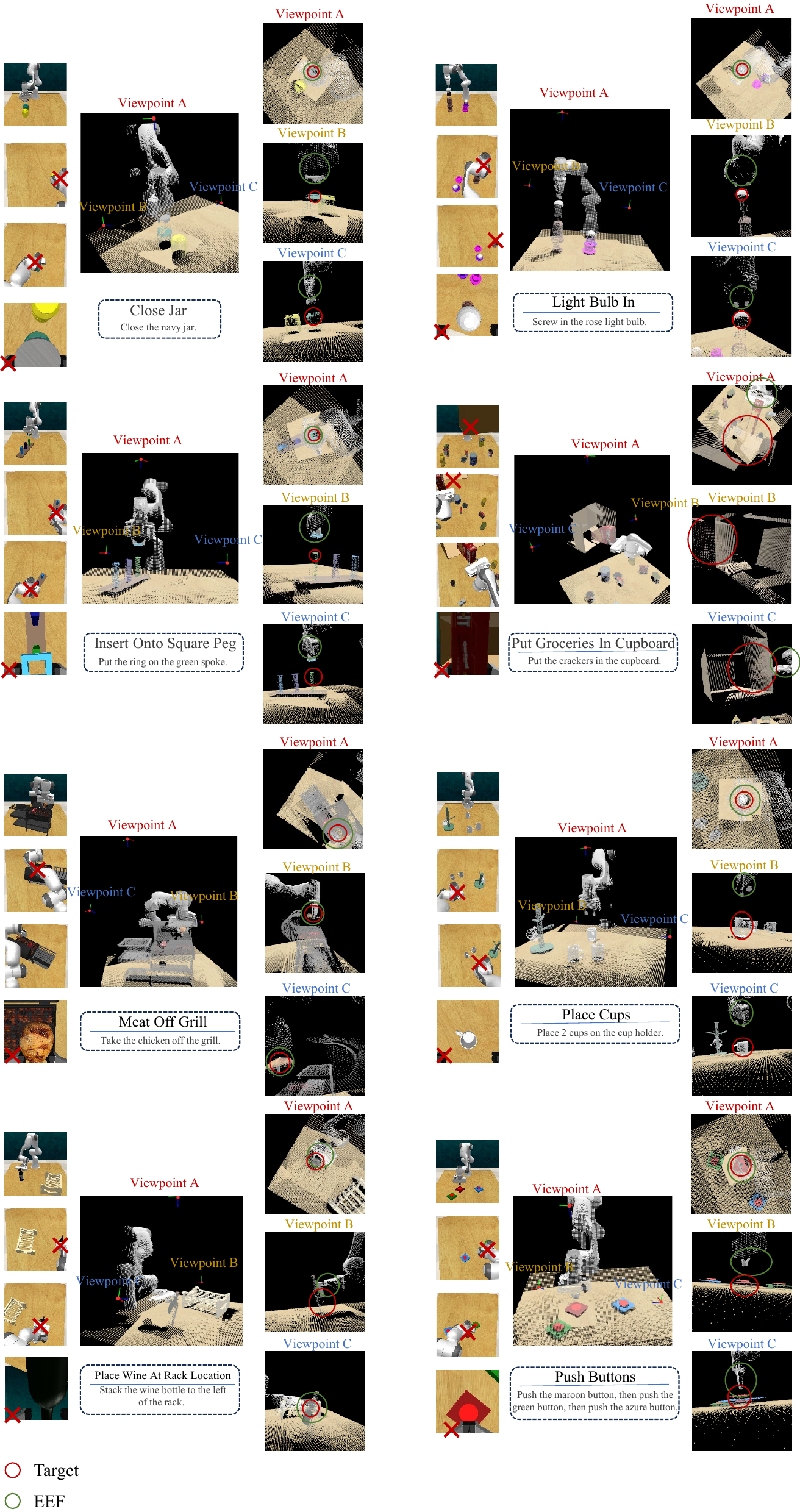}
    \caption{The dynamic multi-view re-rendering visualization results of the \textbf{Light Bulb In}, \textbf{Close Jar}, \textbf{Insert Onto Square Peg}, \textbf{Put Groceries In Cupboard}, \textbf{Grill}, \textbf{Place Cups}, \textbf{Place Wine At Rack Location}, and \textbf{Push Buttons} tasks.}
    \label{fig:rendering-results-1}
\end{figure*}

\begin{figure*}[t]
    \centering
    \includegraphics[width=0.7\textwidth]{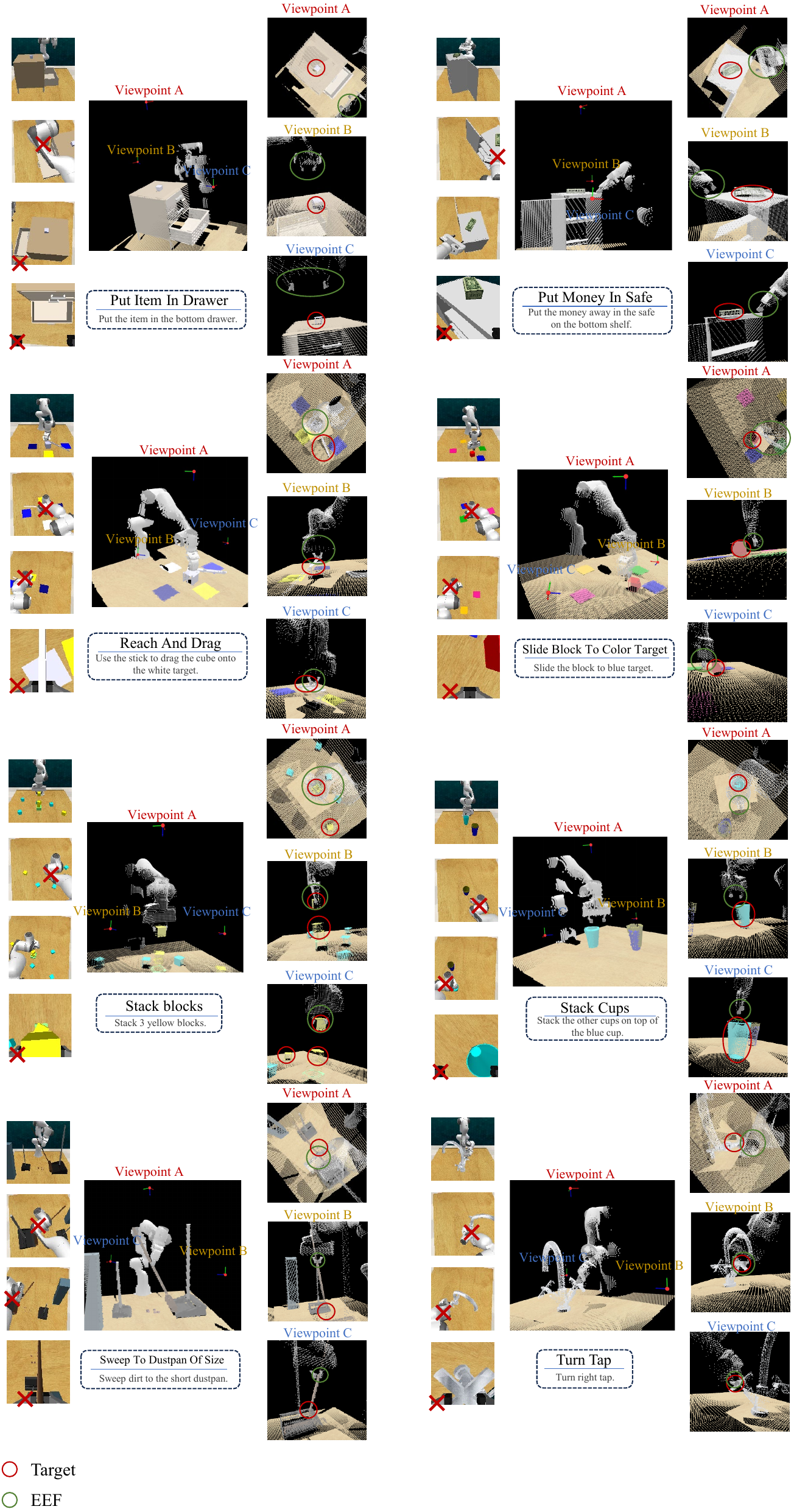}
    \caption{The dynamic multi-view re-rendering visualization results of the \textbf{Put Item In Drawer}, \textbf{Put Money In Safe}, \textbf{Reach And Drag}, \textbf{Slide Block To Color Target}, \textbf{Stack Blocks}, \textbf{Stack Cups}, \textbf{Sweep To Dustpan Of Size}, and \textbf{Turn Tap} tasks.}
    \label{fig:rendering-results-2}
\end{figure*}

To further illustrate the operational principles of our \ModelAbbr\ framework and its performance on RLBench, we visualize multiple dynamic virtual viewpoints generated by MVEP in 3D space at intermediate execution steps during inference, along with their corresponding rendered 2D images for each scene, as shown in Fig.~\ref{fig:rendering-results-1} and Fig.~\ref{fig:rendering-results-2}. The rendered imagery clearly captures both the end-effector and target objects, enabling the action model to make more precise motion predictions based on these perceptual cues. This enhancement directly contributes to improved task success rates across multiple manipulation scenarios. \ModelAbbr\ effectively translates visual completeness into manipulation success, validating that dynamic ``seeing" fundamentally underpins robust ``acting."

\end{document}